\documentclass[10pt,twocolumn,letterpaper]{article}

\usepackage[pagenumbers]{cvpr}

\usepackage{array}
\usepackage{pifont}
\usepackage{xurl}

\usepackage{amsmath,amsfonts,bm}

\def\eqref#1{equation~\ref{#1}}

\def\1{\bm{1}}

\DeclareMathAlphabet{\mathsfit}{\encodingdefault}{\sfdefault}{m}{sl}
\SetMathAlphabet{\mathsfit}{bold}{\encodingdefault}{\sfdefault}{bx}{n}

\newcommand{\R}{\mathbb{R}}

\makeatletter
\@ifpackageloaded{lineno}{%
  \renewcommand{\makeLineNumberOdd}{%
    \if@firstcolumn\makeLineNumberLeft\else\makeLineNumberRight\fi}
  
}{}
\makeatother

\definecolor{cvprblue}{rgb}{0.21,0.49,0.74}
\usepackage[pagebackref,breaklinks,colorlinks,allcolors=cvprblue]{hyperref}

\def\confName{CVPR}
\def\confYear{2027}

\title{CIALLO: Learning Shared Intent Representations for Cross-Embodiment Manipulation}
\author{\begin{minipage}{0.96\textwidth}\centering\normalsize
{\bfseries
\mbox{Choo Sin Wai\textsuperscript{*}},
\mbox{Dake Zhong\textsuperscript{*}},
\mbox{Jingye Zhang\textsuperscript{*}},
\mbox{Zhifang Liu},
\mbox{Siyao Xiao},
\mbox{Zirong Ding},
\mbox{Yuhong Zhang},
\mbox{Kangjie Chen},
\mbox{Dongbin Zhang},
\mbox{Guorui Song},
\mbox{Guocun Wang},
\mbox{Peng Jiao},
\mbox{Haoqian Wang\textsuperscript{\(\dagger\)}}}\\[0.4ex]
Tsinghua University\\
\texttt{caosw25@mails.tsinghua.edu.cn}\\[0.4ex]
{\footnotesize\textsuperscript{*}Equal contribution.\quad
\textsuperscript{\(\dagger\)}Corresponding author.}
\end{minipage}}

\begin{document}
\maketitle

\begin{abstract}
Robot demonstrations contain reusable manipulation knowledge, but transferring this knowledge across embodiments requires representations that bridge differences in control interfaces. Action reconstruction alone does not ensure that similar behaviors across robots share a representation. To address this gap, we introduce CIALLO, an action codec that aligns actions, visual changes, and observation-grounded language in a shared space of short-horizon behavioral intent. An embodiment-conditioned decoder maps this intent back to robot-specific actions, allowing a shared representation to support different control interfaces. After pretraining on eight embodiments, CIALLO adapts to a new robot from one recording of at most 10 seconds by updating only its embodiment embeddings. The calibrated codec then remains frozen while a policy learns to predict intent from observations and instructions. We instantiate this policy with Qwen3.5-4B to form Qwen-CIALLO, whose backbone requires no large-scale robot-action pretraining before policy finetuning. With 20 demonstrations per task, Qwen-CIALLO achieves 76.7\% average success across real SO-101 and single-arm Piper robots, compared with 70.0\% for fully finetuned $\pi_0$. In simulation, we additionally finetune the action decoder, achieving 97.3\% success on LIBERO and 52.7\% on RoboTwin 2.0. The LIBERO policy also achieves 80.8\% on LIBERO-Plus without further finetuning. Together, these results support a shared action codec as a transferable control interface that separates embodiment calibration from task learning.
\end{abstract}

\begin{figure*}[t]
\centering
\includegraphics[width=\textwidth]{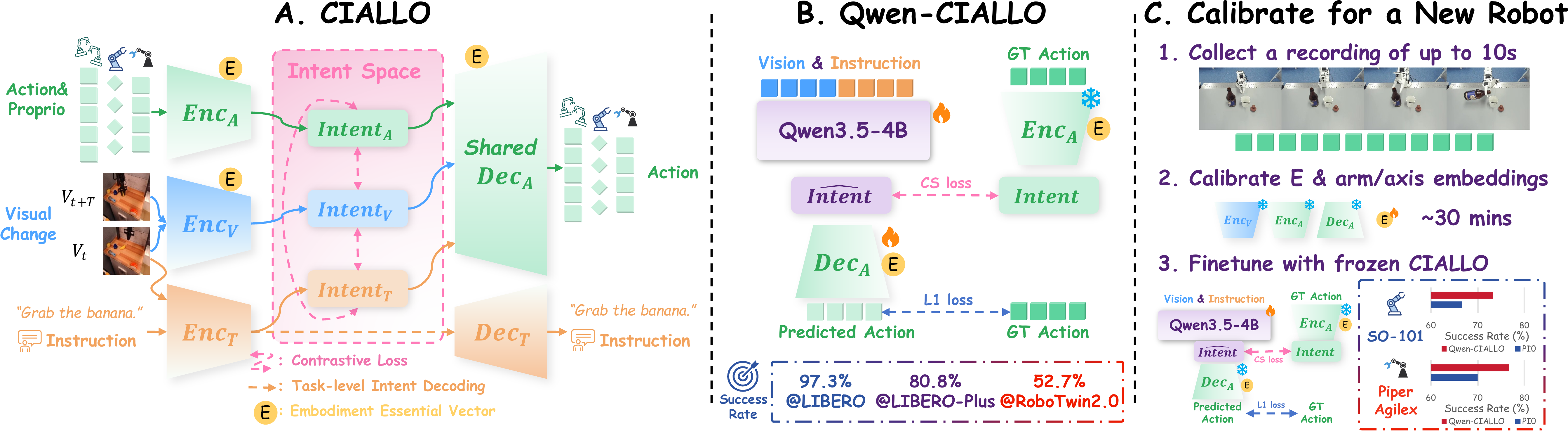}
\caption{\textbf{Overview of CIALLO.} \textbf{(A)} During pretraining, three encoders map an action chunk with proprioception, a pair of current and future frames, and an instruction grounded in the current frame into a shared intent space. We align their outputs with contrastive losses and train a shared decoder to reconstruct the demonstrated actions from each intent, conditioned on the robot's embodiment parameters. \textbf{(B)} To learn a policy, Qwen-CIALLO predicts intent from observations and instructions, using the frozen action encoder as a teacher and the demonstrated actions to supervise the decoder output. CS denotes cosine similarity loss. \textbf{(C)} To transfer the codec to a new robot, we first calibrate its essential vector and arm and axis embeddings while keeping the shared weights fixed, then train the policy with the calibrated codec frozen.}
\label{fig:teaser}
\end{figure*}

\section{Introduction}

A grasp demonstrated by one robot can teach another what to reach for and how an object should move, even when their joint coordinates and control rates differ. Cross-embodiment learning seeks to reuse experience from existing robots to provide control knowledge for a new body~\citep{oxe2023rtx,ghosh2024octo,liu2025rdt}. Pretrained vision-language models supply broad semantic knowledge~\citep{kim2024openvla,Black2024Pi0}, but grounding that knowledge in an unfamiliar action interface still requires substantial amounts of robot demonstration data. A reusable action codec could make these demonstrations more useful for task learning by reducing the need to adapt the underlying control representation.

Recorded actions entangle behavior with the coordinates and control conventions used to execute it. Similar motions can be expressed as end effector displacements or joint angles, with different dimensions, numbers of arms, and temporal resolutions. Shared action layouts and embodiment-specific modules allow policies to consume heterogeneous data~\citep{ghosh2024octo,liu2025rdt,wang2024hpt}. Yet action reconstruction alone, even with embodiment-conditioned encoding and decoding, does not require similar behaviors across robots to have similar representations. Visual changes and grounded language connect different control sequences that produce similar visual effects or serve compatible task goals. We therefore align action representations through these shared cues to learn a codec that transfers across embodiments with only brief calibration.

We introduce CIALLO (\textbf{C}ross-embodiment \textbf{I}ntent \textbf{All} In \textbf{O}ne Space), an action codec for transferring control knowledge across robots. CIALLO separates a shared representation of short-horizon behavior, \emph{intent}, from the body and control schema used to realize it. Building on latent action learning and action--vision alignment~\citep{ye2025lapa,bu2025univla,zhang2026clap,chen2026unit}, CIALLO aligns actions, visual changes, and observation-grounded language in one space. The visual and text branches anchor actions to their effects and task context, encouraging nearby representations for compatible behaviors across embodiments. A shared decoder reconstructs actions from all three representations, preserving the control information needed to execute them (Figure~\ref{fig:teaser}A). An essential vector $E$, arm and axis embeddings, and execution times supply the embodiment-specific context. The same codec weights therefore support different action dimensions, control rates, and single- or dual-arm configurations.

The aligned action codec then provides the interface for policy learning. From one recording of at most 10 seconds, we fit only the new robot's essential vector and arm and axis embeddings, keeping the shared network fixed. We freeze the calibrated codec and train a policy to predict intent from observations and instructions. The action encoder and decoder supervise this prediction (Figure~\ref{fig:teaser}B--C). Calibration establishes how the codec realizes behavior on the new body, while task demonstrations teach the policy which intent to produce. We pair the codec with Qwen3.5-4B~\citep{qwen35blog} through StarVLA~\citep{starvla2026}, forming Qwen-CIALLO without large-scale robot-action pretraining of the Qwen backbone.

We pretrain CIALLO on eight embodiments and evaluate reuse on real SO-101 and single-arm Piper configurations absent from pretraining (Piper pretraining uses a simulated dual-arm configuration). Calibration takes approximately 30 minutes per robot, after which the codec remains frozen. With 20 demonstrations per task, Qwen-CIALLO achieves 76.7\% average success, compared with 70.0\% for fully finetuned $\pi_0$. In simulation, where the action decoder is also finetuned, Qwen-CIALLO reaches 97.3\% on LIBERO after 10k policy steps and 80.8\% on LIBERO-Plus without further training. One policy trained jointly on RoboTwin 2.0 achieves 52.7\% after 40k steps. These evaluations test both adaptation to new robots and policy learning across tasks with the shared action codec.

\section{Related Work}
\label{sec:related}

\subsection{Vision-Language-Action Models and World Models}
\label{sec:paradigm}
Language-conditioned robot policies map observations and instructions to actions~\citep{brohan2022rt1,ghosh2024octo}. Vision-language-action (VLA) models often build on pretrained vision-language backbones~\citep{kim2024openvla,Black2024Pi0,intelligence2025pi05,bjorck2025gr00t}. World models learn dynamics for behavior learning~\citep{hafner2019dreamer} or controllable video generation~\citep{bruce2024genie,liao2025genie,guo2025ctrlworld}. Video-action models jointly learn prediction and control~\citep{zhou2026tau0wm}, while JEPA-VLA transfers frozen video-predictive features into a policy~\citep{miao2026jepavla}. Fast-WAM studies the value of predictive representation learning without test-time future imagination~\citep{yuan2026fastwam}. CIALLO focuses on a shared intent interface with explicit embodiment-conditioned encoding and decoding.

\subsection{Cross-Embodiment Policy Learning}
\label{sec:crossemb}
Pooling robot demonstrations~\citep{oxe2023rtx} requires handling different action dimensions, semantics, and control rates. Approaches include padded unified action spaces~\citep{liu2025rdt}, embodiment- or action-space-specific modules~\citep{ghosh2024octo,bjorck2025gr00t,doshi2024crossformer}, and embodiment-specific stems feeding a shared trunk~\citep{wang2024hpt}. CIALLO conditions shared encoding and decoding networks on an essential vector and control schema. Adaptation fits the new robot's essential vector and arm and axis embeddings, then freezes the calibrated codec during policy learning.

\subsection{Aligning Actions with Visual Change}
\label{sec:align}
Inverse dynamics models recover actions from observation pairs, enabling video labeling from limited action supervision~\citep{baker2022vpt}. Latent-action methods learn a bottleneck between frames without action labels~\citep{schmidt2024lapo,bruce2024genie,ye2025lapa}. UniVLA learns task-centered latent actions for transfer across embodiments~\citep{bu2025univla}. CLAP uses contrastive action--vision alignment~\citep{zhang2026clap}, while UniT uses cross-reconstruction to unify visual and kinematic representations~\citep{chen2026unit}. CIALLO aligns actions, visual change, and observation-grounded language, with a shared action decoder supervising all three paths under the target embodiment context.

\subsection{Action Representations}
\label{sec:actionrep}
Diffusion and flow-matching heads model action distributions through iterative sampling~\citep{chi2023diffusionpolicy,liu2025rdt,Black2024Pi0}. Efficient alternatives include direct regression~\citep{kim2025oft}. Compact policies such as TinyVLA reduce model cost while retaining diffusion decoding~\citep{wen2024tinyvla}. Discrete representations include per-dimension bins~\citep{brohan2022rt1}, frequency-domain action tokens~\citep{pertsch2025fast}, and hierarchical vector quantization~\citep{lee2024vqbet}. Other approaches use spatially structured grids or heatmaps~\citep{james2021c2f,shridhar2022peract,goyal2023rvt,yang2026actionmap}. Drawing on SURE's use of transposed convolutions across neighboring action bins~\citep{zhang2026surespatiotemporaluncertaintyrepresentationefficient}, CIALLO decodes shared intent into continuous actions through an embodiment-conditioned head with a softmax expectation readout.

\newcommand{\actionfig}{%
    \begin{figure*}[t!]
        \centering
        \includegraphics[width=\textwidth]{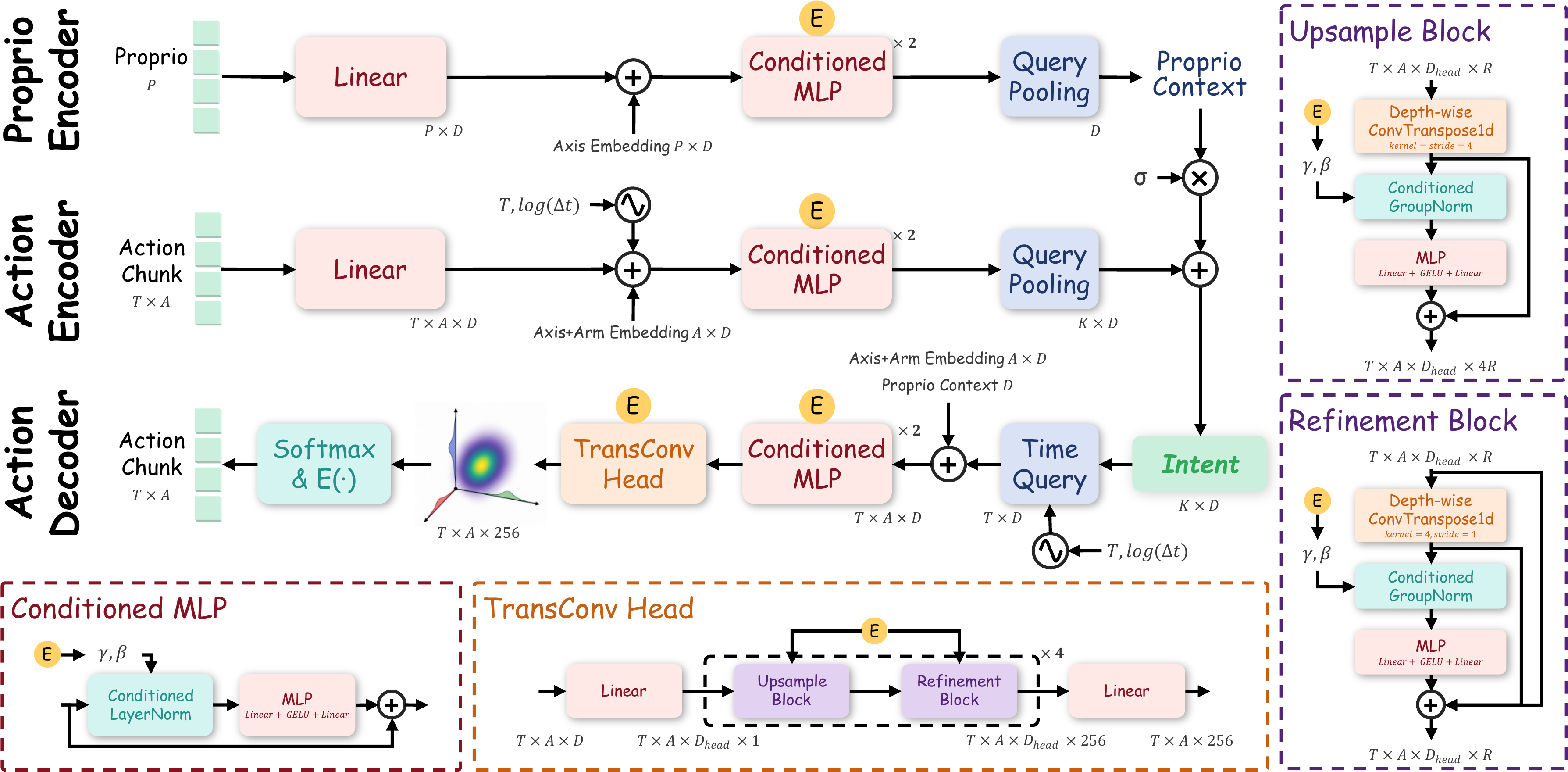}
        \caption{\textbf{The action codec.} The action encoder (middle) uses the essential vector $E$ to map control signals into shared intent, incorporating context from the proprioceptive encoder (top). The decoder (bottom) combines this intent with the target action schema and current proprioception, remodulates the result for the target embodiment, and uses the TransConv Head to predict each control scalar on a shared basis of $256$ ordered bins. The decoded action is the softmax expectation over the bin centers.}
        \label{fig:action}
    \end{figure*}
}

\newcommand{\visiontextfig}{%
    \begin{figure*}[t!]
        \centering
        \includegraphics[width=\textwidth]{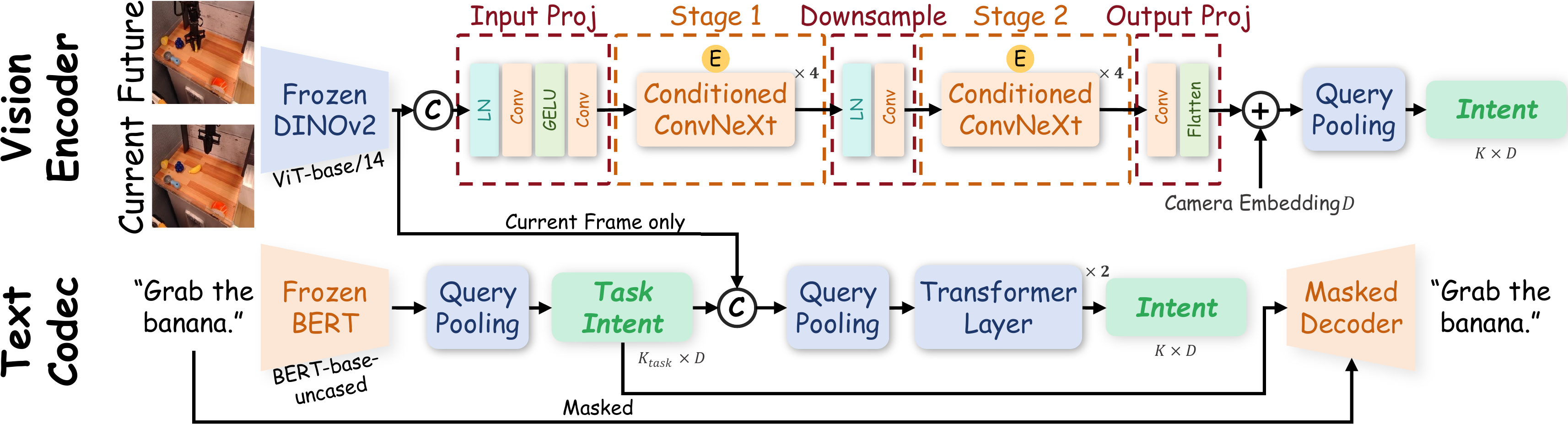}
        \caption{\textbf{Vision and text paths.} The vision encoder compares current and future observations with a shared frozen DINOv2 backbone and an $E$-conditioned local interaction encoder, then pools valid tokens from multiple cameras into chunk-level intent. The text codec compresses the instruction into a task-level intent, grounds it using features reused from the current frame to form the same $K\times D$ intent interface, and reconstructs masked instruction tokens from the task-level intent.}
        \label{fig:vision-text}
    \end{figure*}
}

\section{Method}
\label{sec:method}

\actionfig
\visiontextfig

\subsection{Shared Intent and Embodiment Parameterization}
\label{sec:intent}
A training example contains an instruction, RGB observations, proprioception, and an action chunk $a_{t:t+T-1} \in \R^{T \times A}$ together with its execution times. The chunk length, control rate, and semantics of the $A$ action axes vary across datasets. CIALLO therefore represents a chunk as a masked set of timestamped control scalars rather than assigning semantic meaning to the position of a value in one global action vector. It compresses this set into an \emph{intent} $z \in \R^{K \times D}$, with $K=2$ tokens of width $D=512$ in our experiments. The compact bottleneck provides the same interface for every embodiment, while the encoders and decoder receive the body and control context separately. The text branch also maintains a four-token \emph{task-level intent} that summarizes the instruction. Combining it with the current observation produces a chunk-level intent.

CIALLO makes the complementary, body-specific part of the factorization explicit. Each robot is assigned an \emph{essential vector} $E \in \R^n$, with $n=128$ in practice. Axis embeddings identify whether a scalar controls an end effector component, a joint coordinate, or a gripper command. Arm embeddings further specify the manipulator to which the scalar belongs. Together, these codes provide the information needed to translate between a robot's native signals and the shared intent space. They also provide the parameters used for adaptation. For an unseen robot, we estimate a new $E$ together with its arm and axis embeddings while keeping the shared network fixed, as illustrated in Figure~\ref{fig:teaser}C.

The three encoders produce representations of the same demonstrated chunk,
\begin{equation}
    \begin{aligned}
        z_a &= q_a(a,p,E), \\
        z_v &= q_v(o_t,o_{t+T-1},E), \\
        z_t &= q_t(l,o_t),
    \end{aligned}
    \label{eq:intent-encoders}
\end{equation}
where $p$, $o$, and $l$ denote proprioception, RGB observations, and the instruction. Execution times, axis and arm codes, and validity masks are omitted here for brevity. A shared action decoder reconstructs the chunk from each representation,
\begin{equation}
    \hat a^{m}=p_a(z_m,p,E),\qquad m\in\{a,v,t\}.
    \label{eq:pa}
\end{equation}
Contrastive alignment encourages agreement across the three representations, while reconstruction requires each to retain control information. The text path uses the current observation to ground the instruction and receives neither $E$ nor proprioception. The four-token task-level intent is internal to this path and is distinct from its grounded output $z_t$.

We inject $E$ through conditional normalization~\citep{huang2017adain,perez2018film,peebles2023dit}. A learned projection produces feature-wise scales and shifts $(\gamma,\beta)$, which transform a normalized activation as
\begin{equation}
    \operatorname{CondNorm}(x;E)
    = \operatorname{Norm}(x) \odot (1+\gamma(E)) + \beta(E).
    \label{eq:condnorm}
\end{equation}
The same operation appears on both sides of the intent bottleneck. In the action and visual encoders, it \emph{demodulates} embodiment-specific signals into a shared representation. In the action decoder, it \emph{remodulates} that representation into the coordinates of the target robot. This symmetry provides a common mechanism for the forward and inverse mappings in Eq.~\ref{eq:intent-encoders} and Eq.~\ref{eq:pa}. Architectural and initialization details are given in Supplementary Section~\ref{app:condition}.

\subsection{The Action Codec}
\label{sec:action-codec}
The action codec implements both $q_a$ and $p_a$ with parameters shared across embodiments, as illustrated in Figure~\ref{fig:action}. The encoder represents each valid action scalar by its value, execution time, control axis type, and arm identity. Embodiment-conditioned blocks map these tokens into the common feature space, after which the Query Pooling module compresses the entire chunk into $K$ intent tokens. A separate proprioceptive encoder summarizes the current body state and injects it through a learned gate. The resulting intent is grounded primarily in the executed motion, while proprioception supplies the state and coordinate context needed to interpret that motion.

The decoder reverses this mapping for a requested target interface. The Time Query module reads the intent at the target execution times. It combines the resulting features with target axis and arm codes and current proprioceptive context, then remodulates them under the target essential vector. The temporal schedule and action schema are inputs rather than fixed network properties. One decoder therefore supports different horizons, control rates, action dimensions, and single- or dual-arm layouts. Padding is used only for batching and is removed by masks. It does not impose a shared semantic layout on otherwise unrelated control axes.

For each output scalar, the TransConv Head predicts logits over $256$ uniformly spaced bins in $[-1,1]$, sharing local convolutional kernels along the value axis. A softmax expectation converts these logits into a continuous action, following integral pose regression~\citep{sun2018integral} and DSNT~\citep{nibali2018numerical}. The head is trained with masked $\ell_1$ regression and decodes in one forward pass. Supplementary Section~\ref{app:action} gives the architecture and expectation readout.

\subsection{Vision Encoder}
\label{sec:vision-encoder}

The vision path recovers intent over a short horizon from the change between the current and future observations, as illustrated in Figure~\ref{fig:vision-text}. Both frames from every camera are processed by the same frozen DINOv2 backbone~\citep{oquab2024dinov2}. Freezing this backbone preserves its semantic representation while the interaction encoder learns from heterogeneous robot data. These backbone features remain fixed during embodiment calibration, while the downstream visual representation can change through its dependence on $E$. The paired patch features are concatenated along the channel dimension and processed by a two-stage ConvNeXt interaction encoder conditioned on the essential vector $E$~\citep{liu2022convnext}. This local hierarchy is well suited to the small visual changes produced over a short action chunk. It avoids patch-level global attention and a full cost volume while retaining a sufficient receptive field to relate nearby changes.

The resulting tokens from each available view receive a camera embedding. A validity mask removes absent cameras and padded tokens, so the same encoder handles variable camera counts. Query Pooling summarizes all valid views into the $K\times D$ intent interface shared with the action path. Its learned queries select task-relevant changes across cameras without forming all patch pairs. The DINOv2 features for the current frame are also retained for the text path. Input resolution, stage dimensions, mask definitions, and complexity are detailed in Supplementary Section~\ref{app:vision}.

\subsection{Text Codec}
\label{sec:text-codec}

An instruction usually describes an entire task, whereas chunk-level intent specifies what should happen over the next short horizon. The text path separates task understanding from temporal grounding to bridge these two scopes, as illustrated in Figure~\ref{fig:vision-text}. A frozen BERT encoder~\citep{devlin2019bert} represents the instruction, and Query Pooling compresses its valid token features into a task-level intent. These task-level tokens are fused with visual tokens extracted from the current frame by the frozen vision backbone. A second Query Pooling module extracts $K$ grounded tokens, and two Transformer Layers refine them into a chunk-level intent in $\R^{K\times D}$. This design uses the current observation to identify instruction content relevant to the present chunk, with little additional visual computation.

The task-level intent bottleneck is trained through masked text reconstruction. The Masked Decoder receives an instruction with selected tokens hidden and attends to the task-level intent as its latent memory. The loss is evaluated only at the masked positions. Reconstructing the missing content encourages the task-level intent bottleneck to preserve distinctions expressed by the instruction. The text path intentionally differs from the action and vision paths in its conditioning inputs. It is neither conditioned on $E$ nor given proprioception because language specifies the task independently of the body that performs it. The current observation supplies visual grounding, while alignment across modalities connects the resulting intent to an embodiment-relevant realization. Architectural details and the reconstruction objective are given in Supplementary Section~\ref{app:text}.

\subsection{Supervision}
\label{sec:supervision}

The supervision structure is summarized in Figure~\ref{fig:teaser}A. The action, vision, and text paths each pass their intent through the same action decoder and reconstruct the demonstrated action chunk with masked $\ell_1$ loss. Sharing the decoder makes alignment functional rather than purely geometric. A visual or textual intent must expose the same information that the action codec needs to recover the embodiment-specific controls under the corresponding target context.

The task-level intent is additionally supervised by masked text reconstruction, as described in Section~\ref{sec:text-codec}. Finally, symmetric InfoNCE losses~\citep{radford2021clip} align the three pairs of representations using negatives from the global batch. The complete training objective is
\begin{equation}
    \begin{aligned}
        \mathcal{L} &= \sum_{m\in\{a,v,t\}}\mathcal{L}_{\mathrm{act}}^{m}
        + \mathcal{L}_{\mathrm{text}} + 0.1\mathcal{L}_{a\leftrightarrow v}\\
        &\quad + 0.05\mathcal{L}_{a\leftrightarrow t}
        + 0.05\mathcal{L}_{v\leftrightarrow t}.
    \end{aligned}
    \label{eq:training-objective}
\end{equation}
Here, $a$, $v$, and $t$ index the three paths, $\mathcal{L}_{\mathrm{act}}^{m}$ and $\mathcal{L}_{\mathrm{text}}$ are the masked action and text reconstruction losses, and $\mathcal{L}_{p\leftrightarrow q}$ is symmetric InfoNCE. Unit reconstruction weights keep reconstruction primary, while the smaller alignment weights encourage a shared intent space.

Inspired by FlashAttention and Inf-CL~\citep{dao2022flashattention,cheng2025infcl}, we evaluate exact InfoNCE over the global batch by streaming log-sum-exp blocks and recomputing them during the backward pass. This avoids the full similarity matrix, as detailed in Supplementary Section~\ref{app:contrastive}.

\subsection{Policy Learning and Embodiment Calibration}
\label{sec:policy-transfer}

With the action codec pretrained, we can train a policy to predict intent from observations and instructions, as illustrated in Figure~\ref{fig:teaser}B. Qwen-CIALLO uses a Qwen3.5-4B backbone followed by an intent projector to produce this $K\times D$ representation. The frozen action encoder provides teacher tokens from the demonstrated actions, which supervise the predicted intent through a cosine alignment loss. We also decode the predicted tokens and apply an action regression loss against the demonstrations, training the policy to produce intent that supports accurate control. For the simulation benchmarks, we jointly finetune the backbone, projector, and action decoder. At inference, the backbone predicts intent directly, so the teacher encoder is no longer needed.

When transferring to a new robot, we first fit the embodiment parameters that allow it to use the pretrained codec. As shown in Figure~\ref{fig:teaser}C, we calibrate its essential vector and arm and axis embeddings from a recording of at most 10 seconds while keeping the shared network weights fixed. This optimization takes approximately 30 minutes on one GPU. We then freeze the calibrated codec and train the backbone and projector on demonstrations of the new robot's tasks. Keeping the codec fixed during this second stage lets us evaluate whether it can support policy learning after calibration alone. The real-robot evaluation reports the calibration recording and the policy demonstrations separately to distinguish the data used for these two stages.

\newcommand{\liberoresulttable}{%
    \begin{table*}[tp]
        \centering
        \caption{LIBERO success rates in \%. Results marked with $\dagger$ are averages over three seeds. CLAP-RF and Qwen-CIALLO are trained jointly on all four task suites for 30k and 10k optimization steps, respectively.}
        \label{tab:libero_result}
        \small
        \begin{tabular}{
                    >{\raggedright\arraybackslash}m{6cm}
                    >{\centering\arraybackslash}m{2cm}
                    *{5}{>{\centering\arraybackslash}m{1.15cm}}
                    }
            \toprule
            Method & Steps & Spatial & Object & Goal & Long & Avg. \\
            \midrule
            Diffusion Policy$^\dagger$~\citep{chi2023diffusionpolicy}   & 100--250k     & 78.3              & 92.5              & 68.3              & 50.5              & 72.4              \\
            Octo$^\dagger$~\citep{ghosh2024octo}                        & 100--250k     & 78.9              & 85.7              & 84.6              & 51.1              & 75.1              \\
            OpenVLA$^\dagger$~\citep{kim2024openvla}                    & 50--150k      & 84.7              & 88.4              & 79.2              & 53.7              & 76.5              \\
            OpenVLA-OFT$^\dagger$~\citep{kim2025oft}                    & 50--150k      & 97.6              & 98.4              & 97.9  & \textbf{94.5}     & 97.1  \\
            GR00T-N1$^\dagger$~\citep{bjorck2025gr00t}                  & --            & 94.4              & 97.6              & 93.0              & 90.6              & 93.9              \\
            $\pi_0$$^\dagger$~\citep{Black2024Pi0}                      & --            & 96.8              & 98.8  & 95.8              & 85.2              & 94.2              \\
            StarVLA-PI~\citep{starvla2026}                              & 30k           & \textbf{98.8}     & \textbf{99.6}     & 95.8              & 88.4              & 95.7              \\
            StarVLA-OFT                                                 & 30k           & 97.8              & 98.6              & 96.2              & 93.8              & 96.6              \\
            StarVLA-GR00T                                               & 30k           & 97.8              & 98.8  & 97.4              & 92.0              & 96.5              \\
            CLAP-RF~\citep{zhang2026clap}                               & 30k           & \underline{98.6}  & \underline{99.2}  & \underline{98.0}  & 93.0              & \underline{97.2}  \\
            \midrule
            \textbf{Qwen-CIALLO}                                        & \textbf{10k}  & 98.4  & 98.4              & \textbf{98.2}     & \underline{94.0}  & \textbf{97.3}     \\
            \bottomrule
        \end{tabular}
    \end{table*}
}

\newcommand{\liberoplusresulttable}{%
    \begin{table*}[tp]
        \centering
        \caption{LIBERO-Plus success rates in \%. Total reports overall success across all evaluation tasks for every method. Qwen-CIALLO uses its LIBERO checkpoint without further finetuning.}
        \label{tab:libero_plus_result}
        \small
        \begin{tabular}{
                    >{\arraybackslash}m{4cm}
                    >{\centering\arraybackslash}m{1.0cm}
                    >{\centering\arraybackslash}m{1.0cm}
                    >{\centering\arraybackslash}m{1.3cm}
                    >{\centering\arraybackslash}m{1.0cm}
                    >{\centering\arraybackslash}m{1.6cm}
                    >{\centering\arraybackslash}m{1.0cm}
                    >{\centering\arraybackslash}m{1.0cm}
                    >{\centering\arraybackslash}m{1.0cm}
                    }
            \toprule
            Method                              & Camera            & Robot             & Language          & Light             & Background        & Noise             & Layout            & Total             \\
            \midrule
            $\pi_0$                             & 15.8              & 6.6               & 61.0              & 79.6              & 78.5              & \underline{79.4}  & 70.4              & 53.6              \\
            UniVLA~\citep{bu2025univla}         & 1.8               & 46.2              & 69.6              & 69.0              & 81.0              & 21.2              & 31.9              & 42.9              \\
            OpenVLA                             & 0.8               & 3.5               & 23.0              & 8.1               & 34.8              & 15.2              & 28.5              & 15.6              \\
            OpenVLA-OFT                         & \underline{56.4}  & 31.9              & 79.5              & 88.7              & \underline{93.3}  & 75.8              & \underline{74.2}  & 69.6              \\
            Fast-WAM~\citep{yuan2026fastwam}    & 16.3              & 44.1              & 66.8              & 79.8              & 52.6              & 38.5              & 61.4              & 49.9    \\
            DreamWAM~\citep{yuan2026dreamwam}   & 53.8              & \underline{63.6}  & \textbf{94.8}     & \textbf{96.7}     & 71.6              & 67.2              & \textbf{80.7}     & \underline{74.6} \\
            \midrule
            \textbf{Qwen-CIALLO}                & \textbf{57.5}     & \textbf{66.7}     & \underline{90.5}  & \underline{95.1}  & \textbf{94.6}     & \textbf{88.9}     & \textbf{80.7}     & \textbf{80.8}     \\
            \bottomrule
        \end{tabular}
    \end{table*}
}

\newcommand{\robotwinresulttable}{%
    \begin{table*}[tp]
        \centering
        \caption{RoboTwin 2.0 success rates (\%) under the clean-to-clean protocol. Single Task uses one checkpoint per task, and Co-Train uses one checkpoint jointly trained on all 50 tasks.}
        \label{tab:robotwin_result}
        \small
        \begin{tabular}{
                    >{\arraybackslash}m{3.6cm}
                    >{\centering\arraybackslash}m{2.2cm}
                    >{\centering\arraybackslash}m{1.0cm}
                    >{\arraybackslash}m{3.6cm}
                    >{\centering\arraybackslash}m{2.2cm}
                    >{\centering\arraybackslash}m{1.0cm}
                    }
            \toprule
            Method                      & Protocol      & Avg.      & Method                & Protocol      & Avg.              \\
            \midrule
            Diffusion Policy            & Single Task   & 28.0      & $\pi_0$               & Single Task   & 46.4              \\
            ACT~\citep{zhao2023act}     & Single Task   & 29.7      & StarVLA               & Co-Train      & \underline{46.5}  \\
            RDT~\citep{liu2025rdt}      & Single Task   & 34.5      & \textbf{Qwen-CIALLO}  & Co-Train      & \textbf{52.7}     \\
            \bottomrule
        \end{tabular}
    \end{table*}
}

\newcommand{\realworldresult}{%
    \begin{table*}[tp]
        \centering
        \caption{Real robot task success rates (\%) on SO-101 and single-arm Piper, with 20 training demonstrations and 20 evaluation trials per method and task on each robot. Qwen-CIALLO uses a frozen codec after calibration from one recording per robot, while $\pi_0$ is fully finetuned.}
        \label{tab:real_robot_results}
        \small
        \begin{tabular}{
                    >{\raggedright\arraybackslash}m{2.15cm}
                    *{6}{>{\centering\arraybackslash}m{1.6cm}}
                    >{\centering\arraybackslash}m{1.0cm}
                    }
            \toprule
            & \multicolumn{3}{c}{SO-101} & \multicolumn{3}{c}{Piper} & \\
            \cmidrule(lr){2-4}\cmidrule(lr){5-7}
            Method      & Water bottle & Pour beer  & Wooden fish   & Water bottle  & Pour beer     & Wooden fish   & Avg.      \\
            \midrule
            Qwen-CIALLO & 85 & 80 & 70 & 90 & 95 & 40 & 76.7 \\
            $\pi_0$     & 90 & 80 & 35 & 100 & 95 & 20 & 70.0 \\
            \bottomrule
        \end{tabular}
    \end{table*}
}

\section{Experiments}
\label{sec:experiments}

We first evaluate whether the three intent paths align and retain information needed by the shared action decoder. We then evaluate policies trained through this interface on LIBERO and RoboTwin 2.0, with LIBERO-Plus testing the LIBERO policy under distribution shifts. Finally, we assess adaptation to real robots by calibrating the codec and keeping it frozen during policy training.

\subsection{CIALLO Pretraining}
\label{sec:pretraining-experiments}

We pretrain CIALLO on 125,434 episodes from eight embodiments, with balanced sampling across embodiments. The data span LIBERO~\citep{liu2023libero}, Bridge and Fractal as used by SimplerEnv~\citep{simpler2024}, and RoboTwin 2.0~\citep{chen2025robotwin2}.

On 2,048 held-out samples, action intent retrieves its paired visual or grounded language intent, and vice versa, with bidirectional top-1 accuracies of 96.68\% and 91.94\%, respectively. Retrieval between vision and grounded language reaches 96.88\%. The shared decoder also reconstructs actions from each modality, with normalized masked MAE of 0.00761 from action, 0.01853 from vision, and 0.02429 from grounded language. Supplementary Section~\ref{app:pretraining-diagnostics} details training, evaluation, and results by embodiment.

\subsection{Contrastive Alignment Ablation}

We vary the contrastive terms while retaining all reconstruction losses. Each model uses eight embodiments, two intent tokens, and the same 2,000-step training budget and validation set. Action--vision alignment alone yields the lowest vision-to-action error (Table~\ref{tab:alignment-ablation}). The full objective yields the lowest action-to-action error, a 7.1\% reduction relative to reconstruction alone, and also improves vision-to-action reconstruction over that baseline.

\begin{table}[h]
    \centering
    \caption{Effect of contrastive alignment on reconstruction. MAE is averaged equally over eight embodiments. A, V, and T denote action, vision, and grounded text.}
    \small
    \label{tab:alignment-ablation}
    \begin{tabular}{lcc}
        \toprule
        Contrastive terms & A$\to$A MAE $\downarrow$ & V$\to$A MAE $\downarrow$ \\
        \midrule
        None              & 0.03917 & 0.05750 \\
        A--V              & 0.04042 & \textbf{0.05465} \\
        A--V, A--T        & 0.03809 & 0.05495 \\
        A--V, A--T, V--T  & \textbf{0.03639} & 0.05614 \\
        \bottomrule
    \end{tabular}
\end{table}

\subsection{Embodiment Conditioning}

If $E$ merely selected a matched encoding and decoding convention for compressing actions, using the same $E$ at both ends should largely preserve reconstruction. We test this prediction on 2,048 held-out samples by substituting another robot's $E$ while keeping the inputs and target action schema fixed. Action-to-action MAE rises from 0.00761 to 0.04506 when we change only the encoder's $E$, and to 0.19124 when we change only the decoder's. Giving both stages the same substitute $E$ lowers the error to 0.16470, still more than 21 times the original error. Agreement between the two stages is therefore insufficient for accurate reconstruction. In this test, $E$ must also match the robot whose actions the codec encodes and decodes. Changing only the encoder's $E$ affects reconstruction less for visual intent than for action intent. Supplementary Table~\ref{tab:wrong-embodiment} reports reconstruction errors for all paths and substitution conditions. This lower sensitivity supports our hypothesis that changes in the observed scene provide cues to intent that different robots can share despite differences in their action spaces.

\subsection{Number of Pretraining Embodiments}

With the training objective and step budget fixed, both action-to-action and vision-to-action MAE decrease on each of the three original embodiments (Panda, WidowX, and Google Robot) as pretraining expands from three to five to eight embodiments. For Panda, action-to-action MAE falls from 0.0807 to 0.0339 to 0.0136. Each expansion also introduces more diverse demonstrations from RoboTwin. We hypothesize that training on more embodiments and a wider range of demonstrations helps the shared codec represent a broader range of actions, which could explain the lower reconstruction errors even on the original three embodiments. Supplementary Table~\ref{tab:embodiment-count} gives per-embodiment results and five-embodiment validation counts.

\subsection{Direct Closed-Loop Control}
\label{sec:ciallo-direct}

For closed-loop control without policy finetuning, we feed the intent produced by CIALLO's pretrained text encoder directly to its pretrained action decoder. This CIALLO-Direct setting reaches 73.4\% success on LIBERO Spatial, 52.2\% on Object, 65.6\% on Goal, and 17.8\% on Long. It establishes a direct control baseline for the pretrained representation before coupling the codec to Qwen3.5.

\liberoresulttable
\liberoplusresulttable

\subsection{Qwen-CIALLO}
\label{sec:qwen-ciallo}

\subsubsection{Implementation Details}

We use Qwen3.5-4B with the intent prediction and action supervision described in Section~\ref{sec:policy-transfer}. For LIBERO and RoboTwin 2.0, the action encoder remains frozen while the backbone, intent projector, and action decoder are finetuned. Supplementary Section~\ref{app:qwen-training} lists hyperparameters.

\subsubsection{LIBERO}

\noindent\textbf{Evaluation protocol.} We evaluate Qwen-CIALLO on the four standard LIBERO task suites, namely LIBERO-Spatial, LIBERO-Object, LIBERO-Goal, and LIBERO-Long. Together, these suites assess spatial reasoning, object-centric generalization, goal understanding, and long-horizon manipulation~\citep{liu2023libero}. Baseline results in Table~\ref{tab:libero_result} come from sources that use different statistical protocols. Results marked with $\dagger$ are averaged over three random seeds, with 500 evaluation trials per suite and 50 trials for each of the 10 tasks. Diffusion Policy, Octo, OpenVLA, and OpenVLA-OFT results are taken from the OpenVLA-OFT evaluation~\citep{kim2025oft}. The GR00T-N1 and $\pi_0$ results are reported by Hume~\citep{song2026hume}, and the StarVLA results come from its official evaluations~\citep{starvla2026}. CLAP-RF results are taken from its reported LIBERO evaluation~\citep{zhang2026clap}, using joint training across all four suites for 30k steps. Both the StarVLA variants and Qwen-CIALLO use single-seed evaluations.

\noindent\textbf{Results.} Qwen-CIALLO attains 97.3\% average success across the four LIBERO suites, the highest average in Table~\ref{tab:libero_result}. It exceeds CLAP-RF by 0.1\%, OpenVLA-OFT by 0.2\%, and all reported StarVLA variants. Both Qwen-CIALLO and CLAP-RF use joint training across all four suites, with 10k and 30k policy finetuning steps, respectively. StarVLA also uses 30k steps. This comparison concerns downstream optimization steps and excludes CIALLO pretraining. It is consistent with the motivation that a pretrained action codec can reduce the burden of learning robot control during policy finetuning, although the comparison does not isolate that mechanism.

\robotwinresulttable
\realworldresult

\subsubsection{LIBERO-Plus}

\noindent\textbf{Evaluation protocol.} LIBERO-Plus extends LIBERO with controlled perturbations to camera viewpoints, robot initial states, language instructions, light conditions, background textures, sensor noise, and object layouts~\citep{fei2026liberoplus}. We evaluate the LIBERO checkpoint without further finetuning. Following the standard protocol, we run one trial per task across the four task suites for each perturbation category. Baseline results in Table~\ref{tab:libero_plus_result} are taken from LIBERO-Plus, except for Fast-WAM and DreamWAM, reported by DreamWAM~\citep{yuan2026dreamwam}.

\noindent\textbf{Results.} The same LIBERO checkpoint achieves 80.8\% overall success, exceeding DreamWAM by 6.2\% (Table~\ref{tab:libero_plus_result}). Qwen-CIALLO has the highest camera, robot, background, and noise scores among the listed methods and ties on layout. This performance is obtained without training specifically for these perturbations. We hypothesize that the short policy finetuning schedule helps preserve the backbone's pretrained capabilities and that intent pretraining across embodiments and environments contributes to scene generalization. The current evaluation measures overall robustness and does not separate these two possible contributions.

\subsubsection{RoboTwin 2.0}

\noindent\textbf{Evaluation protocol.} We evaluate Qwen-CIALLO on RoboTwin 2.0 using the official clean-to-clean protocol~\citep{chen2025robotwin2}. The benchmark contains 50 tasks, each with 50 demonstrations for training. We train a single policy jointly across all tasks. In Table~\ref{tab:robotwin_result}, Single Task methods use one checkpoint per task, while Co-Train methods use a single checkpoint jointly trained across all 50 tasks. All baseline results are taken from the official RoboTwin 2.0 leaderboard.

\noindent\textbf{Results.} After 40k policy finetuning steps, one Qwen-CIALLO checkpoint achieves 52.7\% overall success across all 50 tasks, the highest score in Table~\ref{tab:robotwin_result}. It exceeds the jointly trained StarVLA baseline by 6.2\% and the strongest listed task-specific baseline, $\pi_0$, by 6.3\%. The gains over RDT, ACT, and Diffusion Policy are 18.2\%, 23.0\%, and 24.7\%, respectively. These results show that a single policy trained through the shared intent interface can handle diverse manipulation tasks, outperforming both the jointly trained StarVLA baseline and task-specific policies.

\subsection{Adaptation to New Embodiments}
\label{sec:real-robot}

\noindent\textbf{Evaluation protocol.} We evaluate adaptation on two real robot platforms, SO-101 and a single-arm Piper, using three manipulation tasks with 20 demonstrations per task on each robot. Supplementary Section~\ref{app:real-robot-tasks} provides the task instructions and example observations.

\noindent\textbf{Codec calibration.} From one recording of at most 10 seconds per robot, we fit only its essential vector $E$ and axis and arm embeddings, keeping all shared codec weights frozen (Figure~\ref{fig:teaser}C). The frozen DINOv2 features are independent of $E$, and visual reconstruction is less sensitive to encoder mismatch than action reconstruction (Supplementary Table~\ref{tab:wrong-embodiment}). Aligning the new robot's action intent to this visual reference is designed to largely preserve the pretrained intent space.

\noindent\textbf{Task learning.} We have not pretrained a generalist policy to predict CIALLO intents. With the calibrated codec frozen, we therefore train only the Qwen3.5-4B backbone and intent projector on the 20 demonstrations per task to predict the next intent from the current observation and instruction. We compare against full finetuning of pretrained $\pi_0$~\citep{Black2024Pi0} on the same demonstrations and evaluate each method over 20 trials per task on each robot (Table~\ref{tab:real_robot_results}). Supplementary Section~\ref{app:real-training} gives the optimization settings.

\noindent\textbf{Results.} With 20 demonstrations per task, Qwen-CIALLO achieves 76.7\% average success, compared with 70.0\% for $\pi_0$ (Table~\ref{tab:real_robot_results}). The methods tie on beer pouring on both robots. On water-bottle placement, $\pi_0$ leads by 5 percentage points on SO-101 and 10 on Piper. On the wooden-fish task, Qwen-CIALLO achieves higher success, reaching 70\% versus 35\% on SO-101 and 40\% versus 20\% on Piper. This task requires grasping a thin screwdriver before striking the wooden fish, demanding greater grasp precision. These results compare two routes to cross-embodiment transfer. $\pi_0$ transfers a policy pretrained on large-scale, multi-robot action data and then fully finetuned on the target tasks. Qwen-CIALLO transfers the intent space of a separately pretrained codec and uses the target demonstrations to train only Qwen3.5-4B and its intent projector to select intents for the frozen, calibrated codec.

\section{Conclusion}
\label{sec:conclusion}

We presented CIALLO, an action codec that separates shared behavioral intent from embodiment-specific control. During pretraining, visual changes and grounded language connect actions through shared effects and task context. The aligned action codec then supports policy learning across embodiments. On each of two real robots, one short recording calibrates the codec, which remains frozen during policy learning from 20 demonstrations per task. Together with the simulation results, these findings support separating embodiment calibration from task learning. Future work will align human egocentric videos with humanoid actions in the shared intent space and scale CIALLO pretraining to larger, more diverse datasets. We also aim to pretrain generalist vision-language-action policies from vision-language models using the codec.

\section*{LLM Usage Statement}

We used OpenAI Codex for language editing, manuscript organization, \LaTeX{} formatting, and summarizing experiment records provided by the authors for presentation in the manuscript. We did not use generative AI tools to design the methodology, write or modify experimental code, run experiments, or generate experimental data or results. We reviewed all work produced with AI assistance. The authors checked text produced with AI assistance against the manuscript plan and original experiment records, verified citation metadata against primary sources, and compiled and visually inspected the affected pages. We take responsibility for the final content of this work, including text, claims, or artifacts produced with the aid of generative AI.

{
    \small
    \bibliographystyle{ieeenat_fullname}
    \bibliography{main}
}

\clearpage
\appendix

\section{Model Details}
\label{app:model}

\subsection{Embodiment Conditioning}
\label{app:condition}

Each essential vector is a row of a learned embedding table initialized from $\mathcal{N}(0,n^{-1/2})$. We set $n=128$ in practice. A conditioned residual block takes the form
\begin{equation}
    F_E(x)=x+\operatorname{MLP}\!\left(
    \operatorname{LN}(x)\odot(1+\gamma(E))+\beta(E)\right).
    \label{eq:conditioned-block}
\end{equation}
LayerNorm has no affine parameters. The projection that produces $(\gamma,\beta)$ and the final linear layer of the MLP are zero-initialized, so $F_E$ begins as an identity map. The TransConv Head uses GroupNorm in place of LayerNorm and applies the same $E$-conditioned scale and shift.

\subsection{Action Codec}
\label{app:action}

\subsubsection{Action Normalization and Chunk Construction}

For action axis $j$, the raw command $a_{i,j}$ is normalized with a positive scale $s_j$ defined for each source according to
\begin{equation}
    \tilde a_{i,j}=\operatorname{clip}\!\left(\frac{a_{i,j}}{s_j},-1,1\right),
    \qquad
    \hat a_{i,j}=s_j\hat{\tilde a}_{i,j}.
    \label{eq:action-normalization}
\end{equation}
This zero-preserving transform keeps every valid axis on the shared interval $[-1,1]$. Let $f$ be the native control frequency and $H$ the selected physical horizon. We choose the number of steps as $T=\max(1,\lfloor Hf+0.5\rfloor)$. Table~\ref{tab:action-scaling} lists the values used by the final eight-embodiment mixture. Most sources use a horizon of 0.5 seconds. Bridge and Fractal operate at only 5 Hz and 3 Hz, respectively, so we manually extend their horizons to 1 and 2 seconds to retain 5 and 6 action steps.

\begin{table*}[tp]
    \centering
    \caption{Action normalization and chunk construction. Scale vectors follow the native axis order, and $\times 2$ denotes a vector shared by both arms. RT2 denotes RoboTwin 2.0. The 4 LIBERO suites share one entry.}
    \label{tab:action-scaling}
    \small
    \begin{tabular}{
                >{\arraybackslash}m{3.0cm}
                *{3}{>{\centering\arraybackslash}m{1.0cm}}
                >{\arraybackslash}m{5.7cm}
                }
        \toprule
        Source / embodiment & $f$ (Hz) & $T$ & $H$ (s) & Scale $\mathbf{s}$ \\
        \midrule
        LIBERO / Panda & 20 & 10 & 0.5 & $(1,1,1,0.375,0.375,0.375,1)$ \\
        Bridge / WidowX & 5 & 5 & 1.0 & $(0.05,0.075,0.075,0.15,0.15,0.3,1)$ \\
        Fractal / Google & 3 & 6 & 2.0 & $(0.3,0.25,0.375,0.75,0.6,0.75,1)$ \\
        RT2 / Aloha AgileX & 30 & 15 & 0.5 & $(1.25,3,3,2,1.5,3,1)\times2$ \\
        RT2 / Franka & 30 & 15 & 0.5 & $(2,2,2,3,3,3.75,3,1)\times2$ \\
        RT2 / ARX X5 & 30 & 15 & 0.5 & $(1.5,3,2.5,2,1.5,3,1)\times2$ \\
        RT2 / Piper & 30 & 15 & 0.5 & $(1,3,2.5,2,1.25,3,1)\times2$ \\
        RT2 / UR5 & 30 & 15 & 0.5 & $(2.5,3,2.5,2,3,3,1)\times2$ \\
        \bottomrule
    \end{tabular}
\end{table*}

\subsubsection{Action and Proprioceptive Encoders}

The codec uses token width $D=512$ and eight attention heads. For normalized action $\tilde a_{i,j}$ at time step $i$ and axis $j$, the input token is
\begin{equation}
    x_{i,j}=W_a\tilde a_{i,j}+\psi(\tau_i)
    +e^{\mathrm{axis}}_j+e^{\mathrm{arm}}_j.
    \label{eq:action-token}
\end{equation}
Here, $W_a$ is a learned scalar projection, $\tau_i$ denotes the temporal metadata of step $i$, and $e^{\mathrm{axis}}_j$ and $e^{\mathrm{arm}}_j$ are the axis and arm embeddings. The embedding $\psi$ parameterizes continuous time through the normalized phase $u_i\in[0,1]$ and control period $\Delta t_i$,
\begin{equation}
    \begin{aligned}
        \psi(\tau_i) &= \operatorname{MLP}\!\left(
        [u_i,\log\Delta t_i,\{v_{i,k}\}_{k=0}^{7}]\right),\\
        v_{i,k} &= \left(\sin(2\pi2^ku_i),\cos(2\pi2^ku_i)\right).
    \end{aligned}
    \label{eq:time-features}
\end{equation}
Let $X=\{x_{i,j}\}$ collect all action tokens and let $M_a\in\{0,1\}^{T\times A}$ denote their validity mask. An entry of $M_a$ is one only when the corresponding time step and action axis are both valid. Two Conditioned MLP blocks, each with expansion ratio four, demodulate $X$. The proprioceptive path adds an axis embedding to each projected scalar and applies two Conditioned MLP blocks with the same architecture. Its Query Pooling module uses a single learned query to produce the context vector $c_p$. With $Q=\{q_k\}_{k=1}^{K}$ denoting the $K$ learned action queries, the action path is
\begin{equation}
    \begin{aligned}
        \bar z_a&=\operatorname{QueryPooling}_{Q}\!\left(F_E^{(2)}(X),M_a\right),  \\
        z_a&=\operatorname{LN}\!\left(\bar z_a+\sigma(g)c_p\right).
    \end{aligned}
    \label{eq:action-pooling}
\end{equation}
The operator $\operatorname{QueryPooling}_{Q}(\cdot,M_a)$ performs attention from $Q$ to the valid tokens selected by $M_a$, while $\sigma(g)$ is the learned proprioceptive gate shown in Figure~\ref{fig:action}. The pooling queries are initialized from $\mathcal{N}(0,D^{-1/2})$. We initialize $g=-2$ and drop the complete proprioceptive context with probability 0.1 during training. The embedding tables support 128 action axis types, eight arms, and 64 proprioceptive coordinates.

\subsubsection{Action Decoder and TransConv Head}

During decoding, the Time Query module combines the requested timestamps, encoded by Eq.~\ref{eq:time-features}, with a chunk-level intent $z\in\R^{K\times D}$. It uses two attention blocks to produce one temporal feature per requested step. The resulting features are specialized for each target action axis and remodulated by two Conditioned MLP blocks,
\begin{equation}
    \begin{aligned}
        r_{1:T}&=\operatorname{TimeQuery}\!\left( z,\{\psi(\tau_i)\}_{i=1}^{T}\right),  \\
        y_{i,j}&=G_E^{(2)}\!\left(r_i+e^{\mathrm{axis}}_j +e^{\mathrm{arm}}_j+c_p\right).
    \end{aligned}
    \label{eq:action-query}
\end{equation}
Here, $r_i$ is the temporal feature conditioned on intent and $G_E^{(2)}$ denotes the two Conditioned MLP blocks. Both the temporal and action axis masks are retained throughout this computation.

The TransConv Head first uses a linear layer to lift $y_{i,j}$ to $D_{\mathrm{head}}=128$ channels. It then repeats an Upsample Block followed by a Refinement Block 4 times, expanding the value axis as $1\rightarrow4\rightarrow16\rightarrow64\rightarrow256$. Each Upsample Block applies a depthwise transposed convolution with kernel size and stride 4, followed by $E$-conditioned GroupNorm, GELU, and a residual MLP. Each Refinement Block follows the same structure with a depthwise transposed convolution of kernel size 4 and stride 1. A final linear layer produces the bin logits. Writing $H_E$ for the complete TransConv Head, the projection and soft expectation are
\begin{align}
    \ell_{i,j} &= H_E(y_{i,j})\in\R^{256},
    \notag\\
    p_{i,j,b} &= \operatorname{softmax}_b(\ell_{i,j}), \label{eq:bin-logits}\\
    \hat{\tilde a}_{i,j}
    &=\mathbb{E}_{b\sim p_{i,j}}[c_b]
    =\sum_{b=1}^{256}p_{i,j,b}c_b,
    \quad c_b\in[-1,1]. \label{eq:action-expectation}
\end{align}
The symbols $\ell_{i,j}$ and $p_{i,j,b}$ denote the bin logits and the corresponding softmax probabilities, and $c_b$ is the center of bin $b$. With output validity mask $M\in\{0,1\}^{T\times A}$, the action reconstruction objective is
\begin{equation}
    \mathcal{L}_{\mathrm{act}}=\frac{\sum_{i,j}M_{i,j}
    \left|\hat{\tilde a}_{i,j}-\tilde a_{i,j}\right|}
    {\sum_{i,j}M_{i,j}}.
    \label{eq:action-reconstruction}
\end{equation}
The bin logits are supervised only through this expected action. They form a structured decoding basis rather than a calibrated estimate of action uncertainty. Dropout is set to 0.1 in the attention and residual MLP blocks.

\subsubsection{Connection to Integral Coordinate Regression}
\label{app:expectation-gradient}

Our readout follows an established approach to coordinate regression. Integral pose regression~\citep{sun2018integral} shows that a softmax expectation over a heatmap can be trained with coordinate $\ell_1$ loss alone, without direct heatmap supervision. Expectation-based coordinate readout also appears in LoFTR's fine matching stage~\citep{sun2021loftr}. DSNT~\citep{nibali2018numerical} similarly provides a parameter-free expectation operator that connects convolutional heatmaps to continuous coordinates. These results motivate learning the bin representation indirectly through the decoded action.

The convolutional parameterization supplies a separate design motivation. \citet{nibali2018numerical} argue that spatial parameter sharing helps coordinate models generalize across image locations and that a learned fully connected readout can compromise this property. In CIALLO, the corresponding axis indexes numerical action values, and the TransConv Head reuses local kernels along that axis as it expands the bin resolution. This introduces locality and weight sharing into the output representation. The analogy motivates our architecture, but does not establish the image-translation generalization studied by DSNT for robot actions.

Coordinate supervision does not uniquely determine the intermediate representation. DSNT discusses this ambiguity and explores additional regularization to control heatmap shape~\citep{nibali2018numerical}. Bias-Compensated Integral Regression further identifies decoding bias from the softmax--expectation combination and slower convergence under implicit heatmap supervision in pose estimation~\citep{gu2023bcir}. These findings caution against interpreting a differentiable readout as a guarantee of faithful distribution learning or efficient optimization. CIALLO uses neither heatmap regularization nor bias compensation. We treat its bin weights as a learned decoding basis and evaluate the resulting actions, without assuming a particular distribution shape, number of modes, or calibrated uncertainty.

\begin{figure*}[tp]
    \centering
    \includegraphics[width=\textwidth]{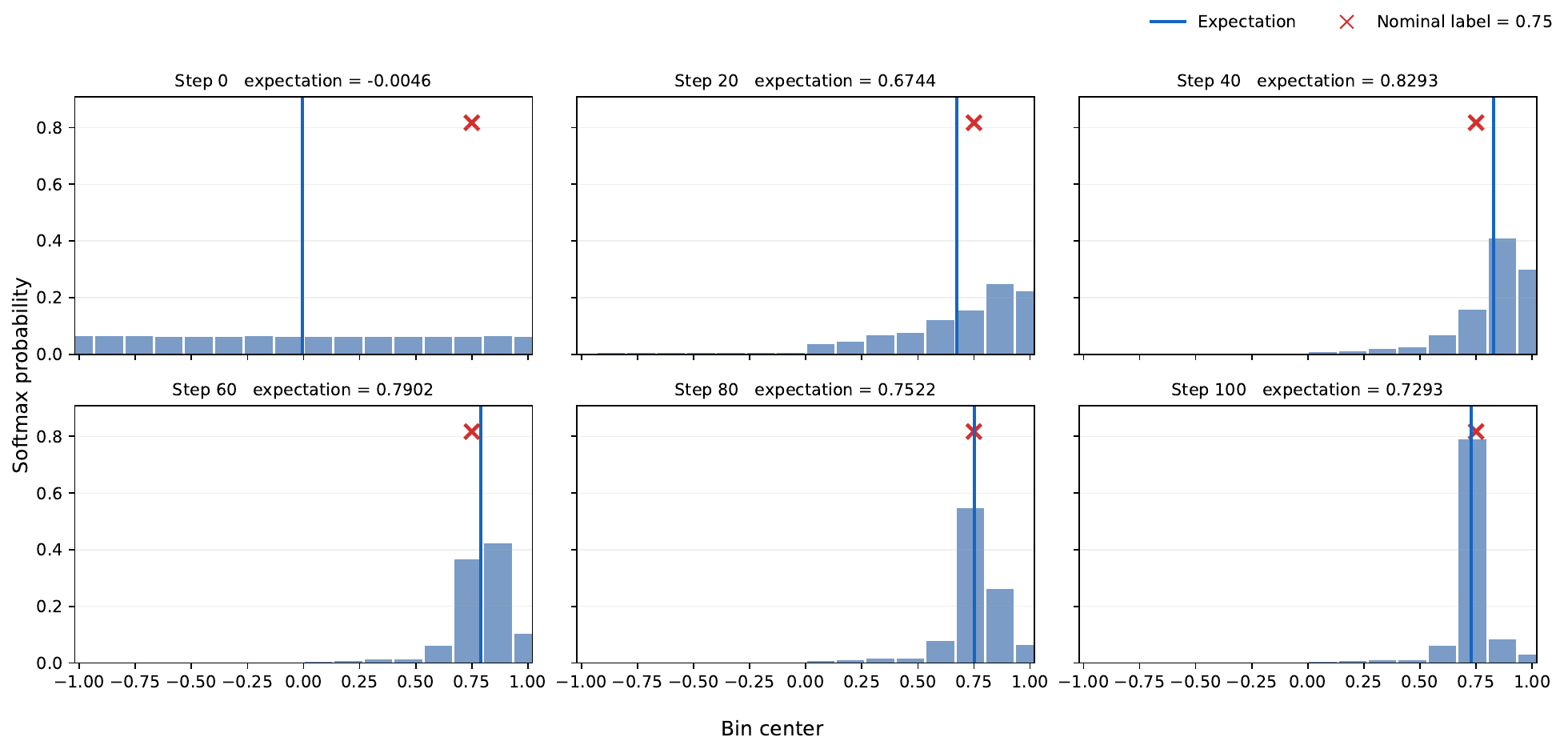}
    \caption{\textbf{Evolution of bin weights under scalar supervision.} Bars show the softmax weights at six training steps. The blue line indicates the predicted scalar, computed as their expectation, and the red cross marks the nominal target of $0.75$. The prediction approaches the target as the weights change, although no loss is applied directly to the bin weights.}
    \label{fig:transconv-demo}
\end{figure*}

Figure~\ref{fig:transconv-demo} illustrates how the bin weights evolve under scalar supervision in a simplified setting. We pass a fixed input through two transposed convolution layers and a shared linear projection to produce logits over 16 ordered bins. We train the network by applying an $\ell_1$ loss to the softmax expectation, using noisy labels around a fixed nominal target of $0.75$.

\subsection{Vision Encoder}
\label{app:vision}

The vision encoder compares the observations at the two endpoints of the same action chunk. For camera $c$, we denote the current frame by $o_c^t$ and the future frame by $o_c^{t+T-1}$, reusing the time index and chunk length $T$ from the action path. Both frames are resized to $252\times252$ and processed by the same frozen DINOv2-base backbone $\Phi$. With a patch size of 14 and the class token removed, the resulting features are
\begin{equation}
    \begin{aligned}
        F_c^t&=\Phi(o_c^t), \\
        F_c^{t+T-1}&=\Phi(o_c^{t+T-1}), \\
        F_c^t,F_c^{t+T-1}&\in\R^{18\times18\times768}.
    \end{aligned}
    \label{eq:visual-backbone}
\end{equation}
The same $\Phi$ is shared across times and cameras, and its parameters receive no gradient updates.

For each camera, we concatenate the two $18\times18$ feature grids along the channel dimension and process them through the modules shown in Figure~\ref{fig:vision-text}. Input Proj applies LayerNorm, convolution, GELU, and convolution to produce features with 256 channels. Stage 1 contains four $E$-conditioned ConvNeXt blocks at this resolution. Downsample uses LayerNorm and a convolution with stride 2 to obtain a $9\times9$ grid with 384 channels, which is processed by the $E$-conditioned blocks in Stage 2. Output Proj maps the result to width $D=512$ and flattens it into 81 tokens per camera.

We support up to 4 camera slots. The 81 tokens from each slot receive its learned camera embedding, and a camera validity mask excludes every token from an absent view. Query Pooling attends from $K=2$ learned queries to the remaining tokens and returns the visual intent in $\R^{K\times D}$. For fixed $K$, this pooling cost grows linearly with the number of visual tokens. The encoder therefore avoids both global self-attention across patches and a pairwise cost volume between the two frames.

\subsection{Text Codec}
\label{app:text}

The frozen BERT encoder produces one feature per instruction token. A learned linear projection maps these features to width $D=512$, while the instruction attention mask removes padding. Query Pooling uses four learned queries to compress the valid features into a four-token task-level intent. The projection and pooling modules are trained while BERT remains frozen.

To ground each chunk, the text path reuses the DINOv2 features $F_c^t$ from the current frame in Eq.~\ref{eq:visual-backbone}, before the interaction encoder for the frame pair. Each camera contributes 324 patch tokens from the current frame. These tokens are projected to width $D$, augmented with camera and visual modality embeddings, and concatenated with the task-level intent after adding a task modality embedding. The combined validity mask keeps all four task-level intent tokens and only the patches from available cameras. A second Query Pooling module reduces this sequence to $K=2$ tokens, and two Transformer Layers produce the chunk-level intent in $\R^{K\times D}$. This path receives neither the essential vector $E$ nor proprioception.

The two-layer Masked Decoder receives the corrupted instruction with learned position embeddings and attends to the task-level intent as memory. Its vocabulary projection is tied to the frozen BERT input embedding, and cross-entropy is averaged only over the selected masked positions. The chunk-level intent is not used as decoder memory. This direct reconstruction path forces the task-level intent to preserve instruction content rather than only the identity of a task class.

\subsection{Memory-Efficient Contrastive Alignment}
\label{app:contrastive}

For modalities $p$ and $q$, we mean-pool each valid chunk-level intent and apply $\ell_2$ normalization to obtain $h_i^p,h_i^q\in\R^D$. A differentiable gather across all devices provides $B$ global pairs. With a learnable inverse temperature $\alpha$ initialized to $1/0.07$, the directional loss is
\begin{equation}
    \begin{aligned}
        \mathcal{L}_{p\rightarrow q} &=\frac{1}{B}\sum_{i=1}^{B} \left[ \log\sum_{j=1}^{B}\exp(s_{ij})-s_{ii} \right],  \\
        s_{ij}&=\alpha\,(h_i^p)^\top h_j^q.
    \end{aligned}
    \label{eq:contrastive-dense}
\end{equation}
Diagonal pairs are positive, and all other samples are negatives. We symmetrize the objective as $\mathcal{L}_{p\leftrightarrow q}=\tfrac{1}{2}(\mathcal{L}_{p\rightarrow q}+\mathcal{L}_{q\rightarrow p})$.

The dense evaluation of Eq.~\ref{eq:contrastive-dense} stores $B^2$ similarities in order to compute $\log\sum_{j=1}^{B}\exp(s_{ij})$ for each row. Following FlashAttention and Inf-CL~\citep{dao2022flashattention,cheng2025infcl}, we implement this log-sum-exp as an exact blockwise streaming computation. We split both axes into blocks of at most $C$ samples and never materialize the full similarity matrix. Let $\mathcal{J}_c$ denote the indices in column block $c$. For an active row $i$, we first compute the largest similarity $\mu_i^{(c)}$ in this block and use it to evaluate the block log-sum-exp without exponentiating unshifted logits.
\begin{equation}
    \begin{aligned}
        \mu_i^{(c)}&=\max_{j\in\mathcal{J}_c}s_{ij},  \\
        \lambda_i^{(c)} &=\mu_i^{(c)}+ \log\sum_{j\in\mathcal{J}_c} \exp\!\left(s_{ij}-\mu_i^{(c)}\right).
    \end{aligned}
    \label{eq:contrastive-block-lse}
\end{equation}
Here $\lambda_i^{(c)}$ is the exact log-sum-exp over column block $c$. We merge it with the running normalization using a second max shift. Starting from $L_i^{(0)}=-\infty$, define
\begin{align}
    \rho_i^{(c)}
    &=\max\!\left(L_i^{(c-1)},\lambda_i^{(c)}\right),
    \label{eq:contrastive-merge-max}\\
    L_i^{(c)}
    &=\rho_i^{(c)}+\log\!\Bigl[
        \exp\!\left(L_i^{(c-1)}-\rho_i^{(c)}\right) \notag\\
    &\qquad +\exp\!\left(\lambda_i^{(c)}-\rho_i^{(c)}\right)\Bigr].
    \label{eq:contrastive-streaming-lse}
\end{align}
Here $\rho_i^{(c)}$ is the larger of the previous running normalization and the current block contribution. After processing blocks $1$ through $c$, the recurrence satisfies $L_i^{(c)}=\log\sum_{d=1}^{c}\sum_{j\in\mathcal{J}_d}\exp(s_{ij})$. The final $L_i$ is therefore exactly the dense log-sum-exp for row $i$, giving the loss $B^{-1}\sum_i(L_i-s_{ii})$. It is accumulated in FP32 while retaining only the current similarity block and one normalization per row.

The backward pass recomputes $L_i$ and revisits the same tiles. Let $H_p^{(r)}$ and $H_q^{(c)}$ be the normalized feature blocks, and define $G_{ij}=[\exp(s_{ij}-L_i)-\mathbb{I}(i=j)]/B$. The feature gradients accumulate as
\begin{equation}
    \begin{aligned}
        \nabla H_p^{(r)} &\mathrel{+}= \alpha G^{(r,c)}H_q^{(c)},  \\
        \nabla H_q^{(c)} &\mathrel{+}= \alpha\bigl(G^{(r,c)}\bigr)^\top H_p^{(r)}.
    \end{aligned}
    \label{eq:contrastive-block-gradients}
\end{equation}
The reverse direction exchanges $p$ and $q$, and the same tiles yield $\nabla\alpha$. This exact reordering reduces memory for the similarity matrix from $B\times B$ to $C\times C$ while retaining quadratic computational complexity.

\section{Pretraining Diagnostics}
\label{app:pretraining-diagnostics}

\noindent\textbf{Data and training setup.} We pretrain CIALLO on 125,434 episodes comprising 1,354 Panda episodes from the four LIBERO suites~\citep{liu2023libero}, 45,541 WidowX episodes from Bridge and 69,339 Google Robot episodes from Fractal as used by SimplerEnv~\citep{simpler2024}, and 9,200 episodes from five RoboTwin 2.0 embodiments~\citep{chen2025robotwin2}. The RoboTwin portion covers Aloha AgileX, Franka Dual, ARX X5, Piper, and UR5. Every source uses only its main camera. We balance sampling across the eight embodiments. The maximum and minimum sample counts differ by 0.30\% of the mean count.

Training runs for 16,000 optimizer steps with a global batch size of 1,024, which processes 16.38 million examples. We use AdamW with weight decay 0.01, a peak learning rate of $3\times10^{-4}$, 400 warmup steps, cosine decay, and BF16 precision. The model has 262.05M parameters, of which 65.99M are trainable because DINOv2 and BERT remain frozen. Training takes 15.7 hours on 8 NVIDIA H20 GPUs, including validation and checkpoint writes.

\subsection{Pretraining Results}
\label{app:pretraining-results}

We evaluate the final checkpoint from one pretraining seed using an episode-level validation split. The evaluation includes 2,048 held-out samples from 1,575 episodes and 512 tasks. Each $2\times512$ intent is flattened in slot order, and exact nearest neighbors are found with raw $\ell_2$ distance. Cross-modal retrieval measures how accurately each representation retrieves its paired demonstration sample. We denote reconstruction from action, visual, and grounded language intent by AA, VA, and TA, respectively. The language representation is the chunk-level intent conditioned on the current observation, distinct from the four-token task-level intent used for instruction reconstruction.

Table~\ref{tab:pretraining-results} reports retrieval within each embodiment and reconstruction through the shared decoder. The higher VA and TA errors for Panda, WidowX, and Google Robot may reflect differences in task coverage, camera setup, control rate, action normalization, and environment. Per-embodiment MAE should therefore not be interpreted as a ranking of representation quality. In the All row, MAE is aggregated over valid action elements, while retrieval searches the complete sample pool.

\begin{table*}[tp]
    \centering
    \caption{Intent retrieval and action reconstruction at the final pretraining checkpoint. Retrieval entries report mean top-1 accuracy across both directions in \%. Per-embodiment retrieval uses candidates from the same embodiment, while the final row uses the complete pool. Reconstruction entries are normalized masked MAE.}
    \label{tab:pretraining-results}
    \begin{tabular}{
                >{\arraybackslash}m{2.1cm}
                >{\centering\arraybackslash}m{1.2cm}
                *{6}{>{\centering\arraybackslash}m{1.2cm}}
                }
        \toprule
        & & \multicolumn{3}{c}{Top-1 retrieval} & \multicolumn{3}{c}{Action MAE} \\
        \cmidrule(lr){3-5}\cmidrule(l){6-8}
        Embodiment & Samples & A$\leftrightarrow$V & A$\leftrightarrow$T & V$\leftrightarrow$T & AA & VA & TA \\
        \midrule
        Panda & 245 & 94.90 & 89.18 & 95.31 & 0.01363 & 0.06924 & 0.08772 \\
        WidowX & 263 & 98.86 & 94.87 & 98.86 & 0.02245 & 0.08469 & 0.11743 \\
        Google Robot & 269 & 98.88 & 93.49 & 99.07 & 0.01851 & 0.09376 & 0.11423 \\
        Aloha AgileX & 259 & 97.68 & 94.98 & 97.88 & 0.00572 & 0.00840 & 0.01098 \\
        Franka Dual & 267 & 95.88 & 89.51 & 95.51 & 0.00691 & 0.01198 & 0.01510 \\
        ARX X5 & 257 & 95.14 & 90.27 & 95.33 & 0.00580 & 0.00919 & 0.01346 \\
        Piper & 245 & 95.71 & 91.02 & 95.31 & 0.00670 & 0.01093 & 0.01580 \\
        UR5 & 243 & 96.30 & 92.18 & 97.53 & 0.00620 & 0.00943 & 0.01201 \\
        \midrule
        \textbf{All} & \textbf{2,048} & \textbf{96.68} & \textbf{91.94} & \textbf{96.88} & \textbf{0.00761} & \textbf{0.01853} & \textbf{0.02429} \\
        \bottomrule
    \end{tabular}
\end{table*}

\begin{table*}[tp]
    \centering
    \caption{AA and VA MAE with increasing numbers of pretraining embodiments. Dashes indicate embodiments absent from the corresponding training setting. Lower values are better. The five-embodiment evaluation uses 819, 857, 799, 814, and 807 validation samples for the first five rows, respectively.}
    \label{tab:embodiment-count}
    \begin{tabular}{
                >{\raggedright\arraybackslash}m{2.4cm}
                *{6}{>{\centering\arraybackslash}m{1.35cm}}
                }
        \toprule
        & \multicolumn{2}{c}{3 embodiments} & \multicolumn{2}{c}{5 embodiments} & \multicolumn{2}{c}{8 embodiments} \\
        \cmidrule(lr){2-3}\cmidrule(lr){4-5}\cmidrule(l){6-7}
        Embodiment & AA & VA & AA & VA & AA & VA \\
        \midrule
        Panda & 0.0807 & 0.1002 & 0.0339 & 0.0766 & 0.0136 & 0.0692 \\
        WidowX & 0.0920 & 0.1107 & 0.0331 & 0.0986 & 0.0225 & 0.0847 \\
        Google Robot & 0.0899 & 0.1129 & 0.0336 & 0.1074 & 0.0185 & 0.0938 \\
        Aloha AgileX & -- & -- & 0.0467 & 0.0493 & 0.0057 & 0.0084 \\
        Franka Dual & -- & -- & 0.0506 & 0.0553 & 0.0069 & 0.0120 \\
        ARX X5 & -- & -- & -- & -- & 0.0058 & 0.0092 \\
        Piper & -- & -- & -- & -- & 0.0067 & 0.0109 \\
        UR5 & -- & -- & -- & -- & 0.0062 & 0.0094 \\
        \bottomrule
    \end{tabular}
\end{table*}

\subsection{Number of Pretraining Embodiments}
\label{app:embodiment-count}

We compare pretraining on three, five, and eight embodiments using the same training objective and number of optimization steps. The three-embodiment setting uses Panda, WidowX, and Google Robot. The five-embodiment setting adds Aloha AgileX and Franka Dual from RoboTwin 2.0, and the eight-embodiment setting further adds ARX X5, Piper, and UR5. Table~\ref{tab:embodiment-count} reports AA and VA reconstruction errors for each setting, with eight-embodiment results taken from Table~\ref{tab:pretraining-results}.

AA and VA MAE decrease for each of the three original embodiments as the number of training embodiments increases from three to five to eight. Both errors also decrease for Aloha AgileX and Franka Dual when training expands from five to eight embodiments. This trend suggests that greater embodiment diversity, together with the addition of RoboTwin data, helps CIALLO learn shared cross-embodiment intent representations.

\subsection{Intent Capacity Ablation}
\label{app:intent-capacity}

We compare intent capacities of $K=1$, $2$, and $4$ tokens, keeping the token width fixed at $D=512$. All other training settings remain fixed, including the full three-modal objective in Eq.~\ref{eq:training-objective}. We evaluate each run after 2,000 optimization steps. Table~\ref{tab:intent-capacity} reports validation losses, macro AA and VA MAE, and peak GPU memory. Macro MAE averages the per-embodiment MAEs equally across all eight embodiments, regardless of action dimensionality or horizon.

\begin{table*}[tp]
    \centering
    \caption{Intent capacity comparison after 2,000 training steps. Lower values are better for all metrics. Losses follow the aggregation used by the validation objective, whereas macro MAE weights embodiments equally. Memory is reported in GiB. Bold values indicate the lowest loss or MAE in each column.}
    \label{tab:intent-capacity}
    \small
    \begin{tabular}{
                >{\centering\arraybackslash}m{0.5cm}
                *{5}{>{\centering\arraybackslash}m{1.32cm}}
                *{2}{>{\centering\arraybackslash}m{1.54cm}}
                }
        \toprule
        & \multicolumn{3}{c}{Validation loss} & \multicolumn{2}{c}{Macro MAE} & \multicolumn{2}{c}{Peak memory (GiB)} \\
        \cmidrule(lr){2-4}\cmidrule(lr){5-6}\cmidrule(l){7-8}
        $K$ & Total & AA & VA & AA & VA & Allocated & Reserved \\
        \midrule
        1 & 0.264283 & 0.024807 & 0.036571 & \textbf{0.035501} & 0.056521 & 37.412 & 38.494 \\
        2 & \textbf{0.261221} & \textbf{0.023484} & \textbf{0.034908} & 0.036392 & \textbf{0.056142} & 37.414 & 38.500 \\
        4 & 0.304822 & 0.033762 & 0.045132 & 0.044961 & 0.065114 & 37.418 & 38.508 \\
        \bottomrule
    \end{tabular}
\end{table*}

Two tokens yield the lowest total, AA, and VA validation losses and the lowest macro VA MAE. One token performs similarly and yields a slightly lower macro AA MAE. Four tokens increase total validation loss by 16.7\% relative to two tokens and increase both macro MAEs. Peak memory is nearly unchanged across capacities, with ranges of 0.006\,GiB for allocated memory and 0.014\,GiB for reserved memory. These results support our choice of $K=2$.

\subsection{Contrastive Alignment Ablation}
\label{app:alignment-ablation}

We compare four combinations of contrastive losses while retaining AA, VA, TA, and masked text reconstruction in every setting. All models are trained on eight embodiments with $K=2$ for 2,000 optimization steps, using the same validation set and otherwise identical training settings. The configurations comprise reconstruction alone, reconstruction with action--vision alignment, additional action--text alignment, and the full objective with vision--text alignment. Table~\ref{tab:alignment-ablation} reports macro AA and VA MAE. The full-model values are taken from the $K=2$ setting in Table~\ref{tab:intent-capacity}.

\subsection{Intent Geometry}
\label{app:intent-geometry}

\begin{figure*}[tp]
    \centering
    \includegraphics[width=0.7\textwidth]{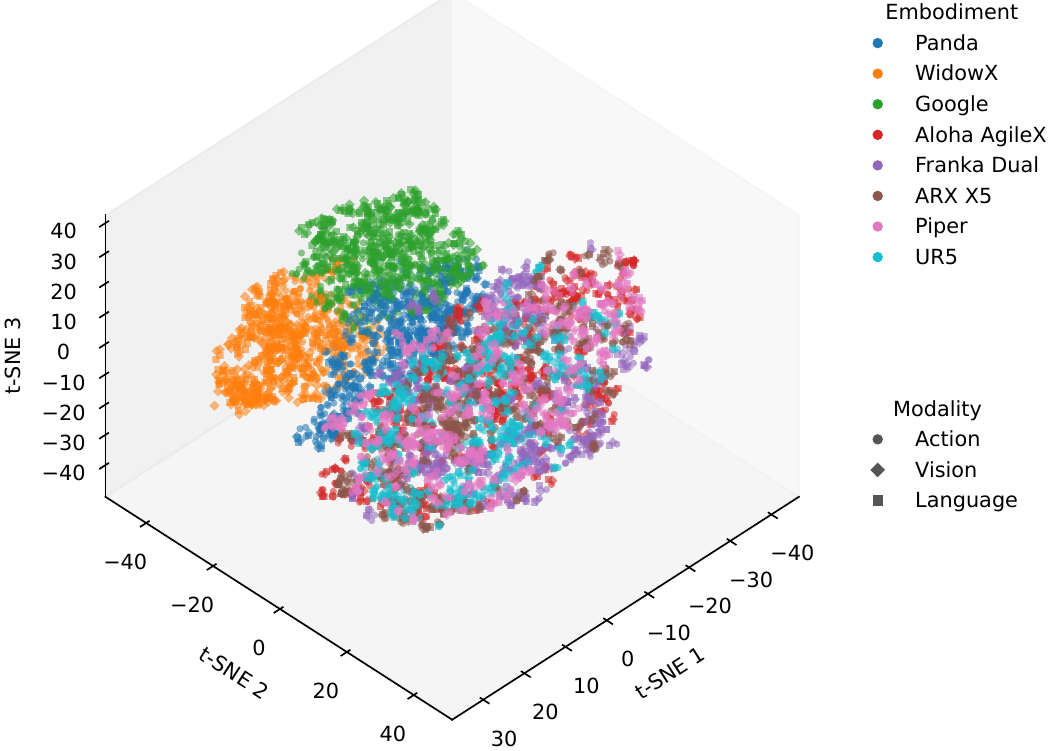}
    \caption{Intent geometry at the final pretraining checkpoint. Color identifies the embodiment and marker shape identifies the input modality. A shared three-dimensional t-SNE embedding is fitted after reducing the flattened intents to 50 dimensions with PCA. The projection emphasizes local neighborhood structure, while global distances should not be interpreted quantitatively.}
    \label{fig:intent-tsne}
\end{figure*}

Figure~\ref{fig:intent-tsne} provides a complementary view of the learned geometry. The five RoboTwin embodiments occupy a broad shared region, consistent with their common task set and similar simulated environments. Panda, WidowX, and Google Robot occupy more distinct, partially overlapping regions, which may reflect differences in their tasks and environments. Within each embodiment, the three modalities exhibit similar local structure. These observations agree with the retrieval results in Table~\ref{tab:pretraining-results}, but the t-SNE axes and global cluster distances have no direct meaning and do not establish strict embodiment invariance.

\subsection{Embodiment Conditioning}
\label{app:embodiment-diagnostic}

We test the functional role of the essential vector on the same 2,048 validation samples used in Section~\ref{sec:pretraining-experiments}. Each sample is assigned the next embodiment in a fixed cycle from Panda through WidowX, Google Robot, Aloha AgileX, Franka Dual, ARX X5, Piper, and UR5 before returning to Panda. We keep the action, observations, instruction, proprioception, timestamps, masks, and target action schema unchanged. The wrong encoder condition replaces $E$ only in the action or visual encoder and its proprioceptive context. The wrong decoder condition replaces $E$ only in the shared action decoder. The all wrong condition applies both replacements. This protocol tests conditioning under a controlled mismatch. It does not retarget an action to another robot.

Table~\ref{tab:wrong-embodiment} shows that the wrong decoder condition raises MAE for all three paths to approximately 0.19--0.21. The wrong action encoder also raises AA MAE from 0.00761 to 0.04506. Replacing the visual encoder condition has a smaller effect on VA, whose MAE rises from 0.01853 to 0.02189. The text path does not consume $E$ or proprioception, so its wrong encoder value is unchanged by construction. These results show that the action encoder and decoder use the essential vector, while the visual path remains sensitive to it to a lesser degree. The errors need not increase monotonically across conditions because simultaneous encoder and decoder mismatches can partially compensate for one another.

\begin{table*}[tp]
    \centering
    \caption{Normalized masked action MAE under essential vector mismatch. Parentheses report the relative increase from the correct condition. Only embodiment conditioning is changed, while every sample retains its original target action schema.}
    \label{tab:wrong-embodiment}
    \begin{tabular}{
                >{\arraybackslash}m{1.2cm}
                >{\arraybackslash}m{1.5cm}
                *{3}{>{\arraybackslash}m{3.0cm}}
                }
        \toprule
        Path & Correct & Wrong encoder & Wrong decoder & All wrong \\
        \midrule
        AA & 0.00761 & 0.04506 (+492.1\%) & 0.19124 (+2413.0\%) & 0.16470 (+2064.3\%) \\
        VA & 0.01853 & 0.02189 (+18.1\%) & 0.20082 (+983.8\%) & 0.21534 (+1062.1\%) \\
        TA & 0.02429 & 0.02429 (+0.0\%) & 0.20684 (+751.5\%) & 0.20684 (+751.5\%) \\
        \bottomrule
    \end{tabular}
\end{table*}

\section{Training Details}
\label{app:training-details}

\subsection{Qwen-CIALLO}
\label{app:qwen-training}

For LIBERO, we train a single Qwen-CIALLO policy jointly on all four task suites for 10,000 optimization steps on 8 NVIDIA H20 GPUs. The per-device batch size is 16. We use AdamW with weight decay $10^{-8}$, 5,000 warmup steps, and cosine annealing. The Qwen3.5-4B backbone, CIALLO action decoder, and intent projector use learning rates of $2\times10^{-5}$, $2\times10^{-6}$, and $1\times10^{-4}$, respectively.

For RoboTwin 2.0, we train a single Qwen-CIALLO policy jointly across all 50 tasks for 40,000 optimization steps on 8 NVIDIA H20 GPUs. The per-device batch size is 4. We otherwise use the LIBERO training configuration, including AdamW with weight decay $10^{-8}$, 5,000 warmup steps, cosine annealing, and learning rates of $2\times10^{-5}$ for the Qwen3.5-4B backbone, $2\times10^{-6}$ for the CIALLO action decoder, and $1\times10^{-4}$ for the intent projector.

\subsection{Adaptation to New Embodiments}
\label{app:real-training}

\noindent\textbf{Calibration settings.} We optimize masked $\ell_1$ action reconstruction from action and visual intents and a contrastive loss between the two intents (Section~\ref{sec:supervision}). The text branch is unused. The 128-dimensional essential vector and the axis and arm embeddings comprise 174.08k parameters. For each robot, we run 5,000 AdamW steps with a batch size of 64 and a learning rate of $10^{-2}$, taking approximately 30 minutes on one NVIDIA H20 GPU. The action chunk size and execution horizon are both 15 steps.

\noindent\textbf{Qwen-CIALLO.} After calibration, we freeze the entire action codec and train the policy for 30,000 steps on 8 NVIDIA H20 GPUs with a per-device batch size of 4. We use AdamW with 2,500 warmup steps and learning rates of $2\times10^{-5}$ for the Qwen3.5-4B backbone and $10^{-4}$ for the intent projector. The action chunk size and execution horizon both remain 15 steps.

\noindent\textbf{$\pi_0$.} We initialize from the pretrained $\pi_0$-base checkpoint and finetune the full model for 30,000 steps on 8 NVIDIA H20 GPUs with a per-device batch size of 4. We use AdamW with a learning rate of $2\times10^{-5}$ and 2,500 warmup steps. The action chunk size and execution horizon are both 50 steps.

\section{Action Head Efficiency}
\label{app:action-head-efficiency}

We compare the direct point-regression head used by OpenVLA-OFT~\citep{kim2025oft}, the flow-matching head of $\pi_{0.5}$~\citep{intelligence2025pi05}, and the CIALLO action decoder. CIALLO uses a distribution-based regression readout that predicts softmax weights over 256 ordered bins and outputs their expectation in a single forward pass (Eq.~\ref{eq:action-expectation}). The bin weights are learned through action regression and serve as a structured decoding basis, rather than a calibrated action distribution.

\noindent\textbf{Benchmark protocol.} All heads are benchmarked on a single NVIDIA H20 GPU in BF16 precision, with batch size 1, an action horizon of 8 steps, and action dimension 7. We report mean latency over 100 inference runs after 20 warmup runs. Table~\ref{tab:action-head-efficiency} reports head parameter counts, FLOPs, and inference latency, excluding the policy backbone. For comparison within the framework used by $\pi_{0.5}$, we also implement the CIALLO decoder in JAX with the same architecture and action readout. This version is labeled ``(JAX)".

\begin{table}[h!]
    \centering
    \caption{Action head efficiency on a single NVIDIA H20 with BF16, batch size 1, an action horizon of 8 steps, and action dimension 7. Measurements exclude the policy backbone. Latency is averaged over 100 inference runs after 20 warmup runs.}
    \label{tab:action-head-efficiency}
    \small
    \begin{tabular}{
                >{\arraybackslash}m{2.3cm}
                *{3}{>{\arraybackslash}m{1.3cm}}
                }
        \toprule
        Method & Params & FLOPs & Latency \\
        \midrule
        OFT~\citep{kim2025oft}                          & 151.12M   & 2.42G     & 0.37ms    \\
        $\pi_{0.5}$ (JAX)~\citep{intelligence2025pi05}  & 430.10M   & 79.13G    & 19.88ms   \\
        \midrule
        CIALLO                                          & 19.89M    & 46.49G    & 4.89ms    \\
        CIALLO (JAX)                                    & 19.89M    & 46.49G    & 1.46ms    \\
        \bottomrule
    \end{tabular}
\end{table}

\noindent\textbf{Results.} CIALLO uses 19.89M parameters, approximately $7.6\times$ fewer than OFT and $21.6\times$ fewer than the $\pi_{0.5}$ action head. Under the same JAX framework, CIALLO decodes an action chunk in 1.46\,ms, compared with 19.88\,ms for $\pi_{0.5}$, a $13.6\times$ speedup. Its 46.49G FLOPs are also below the 79.13G FLOPs of the flow-matching head. The original CIALLO implementation takes 4.89\,ms, showing that the implementation framework materially affects latency even with an unchanged architecture. The single-pass expectation readout avoids the iterative action refinement used by flow matching.

OFT remains the fastest head at 0.37\,ms and uses the fewest FLOPs (2.42G). CIALLO instead combines a smaller parameter count with a structured action-value representation and substantially lower latency than the flow-matching baseline. At the policy level, Qwen-CIALLO achieves higher success than OpenVLA-OFT on LIBERO (97.3\% versus 97.1\%, Table~\ref{tab:libero_result}) and LIBERO-Plus (80.8\% versus 69.6\%, Table~\ref{tab:libero_plus_result}).

\section{Real Robot Tasks}
\label{app:real-robot-tasks}

We use SO-101 and single-arm Piper, each equipped with a gripper, to evaluate manipulation in a tabletop workspace. Demonstrations are recorded at 30\,Hz with a wrist camera and a fixed front camera, and actions are represented as joint angles. Each robot contributes 20 demonstrations for each of the three tasks in Table~\ref{tab:real-robot-instructions}. Figure~\ref{fig:real-robot-examples} shows front camera observations sampled from these demonstrations.

\begin{table}[h!]
    \centering
    \caption{Task instructions used for data collection and evaluation on both robots. The wording is reproduced from the dataset.}
    \label{tab:real-robot-instructions}
    \small
    \setlength{\tabcolsep}{3pt}
    \begin{tabular}{
                >{\arraybackslash}m{1.58cm}
                >{\arraybackslash}m{6.25cm}
                }
        \toprule
        Task & Instruction \\
        \midrule
        Water bottle & Grab the water bottle and place on the black plate. \\
        Pour beer & Pour beer from bottle into the cup. \\
        Wooden fish & Pick up the screwdriver and hit the wooden fish. \\
        \bottomrule
    \end{tabular}
\end{table}

\begin{figure}[htb]
    \centering
    \setlength{\tabcolsep}{1pt}
    \begin{tabular}{ccc}
        \multicolumn{3}{c}{SO-101 / Water bottle} \\
        \includegraphics[width=.32\linewidth]{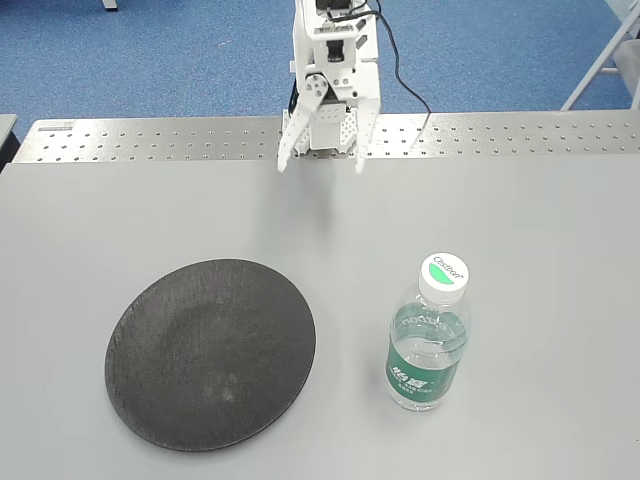} &
        \includegraphics[width=.32\linewidth]{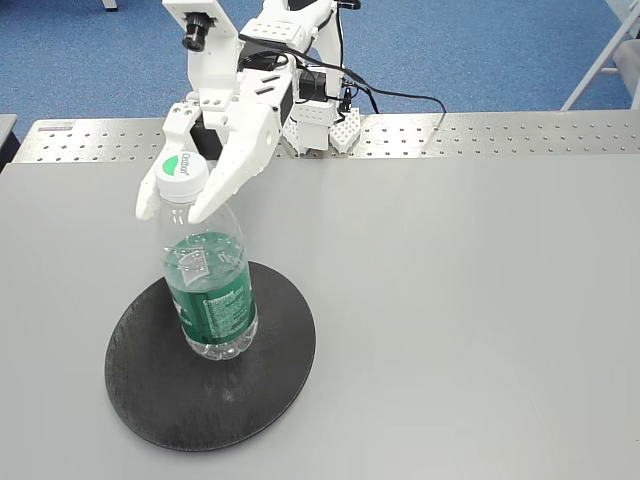} &
        \includegraphics[width=.32\linewidth]{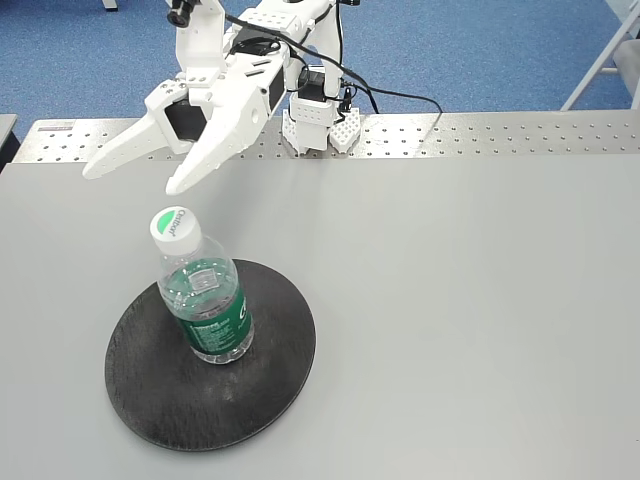} \\
        \multicolumn{3}{c}{SO-101 / Pour beer} \\
        \includegraphics[width=.32\linewidth]{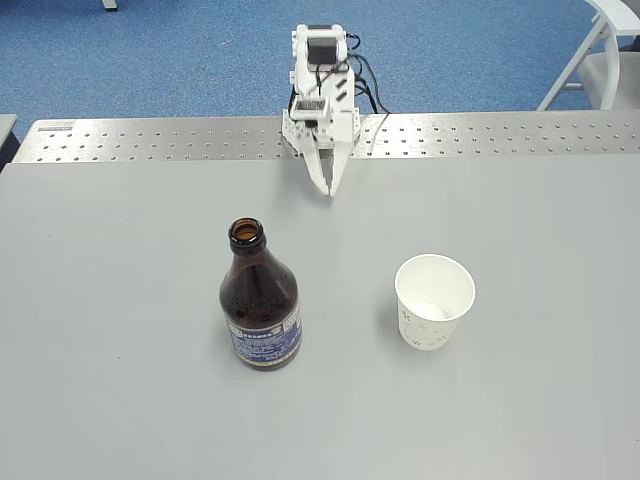} &
        \includegraphics[width=.32\linewidth]{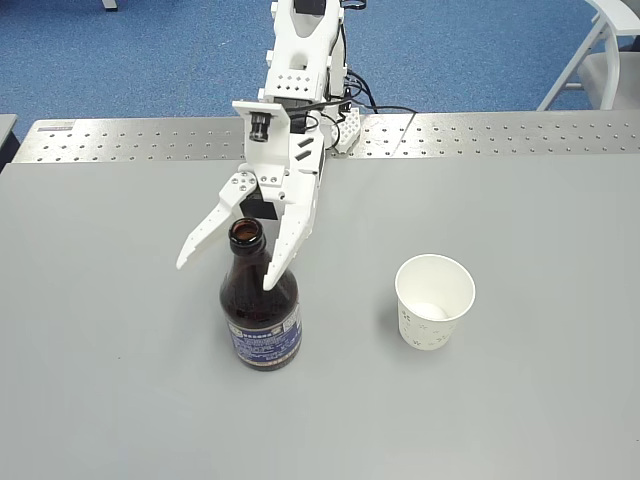} &
        \includegraphics[width=.32\linewidth]{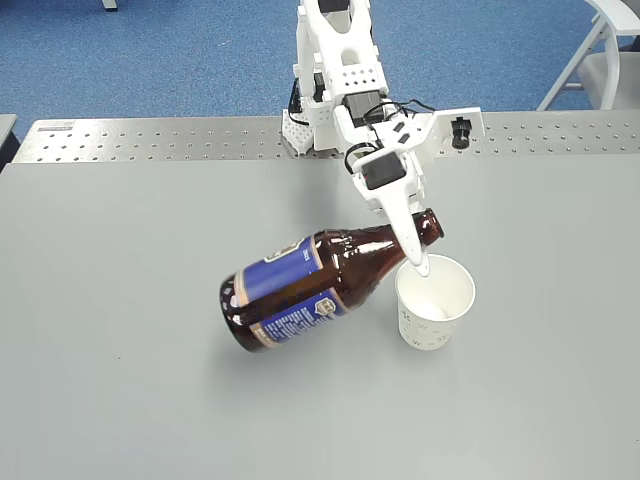} \\
        \multicolumn{3}{c}{SO-101 / Wooden fish} \\
        \includegraphics[width=.32\linewidth]{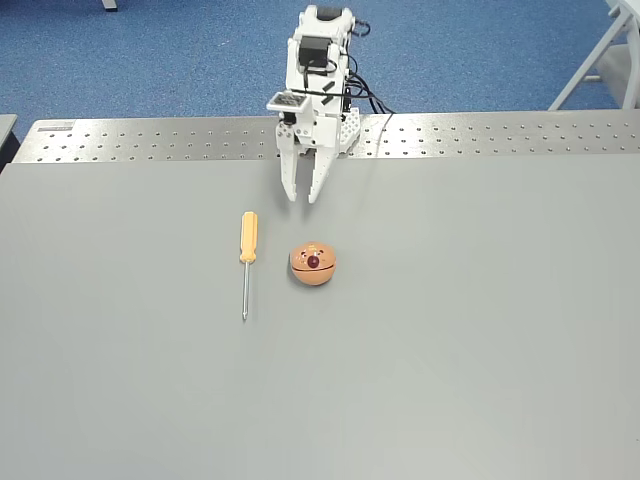} &
        \includegraphics[width=.32\linewidth]{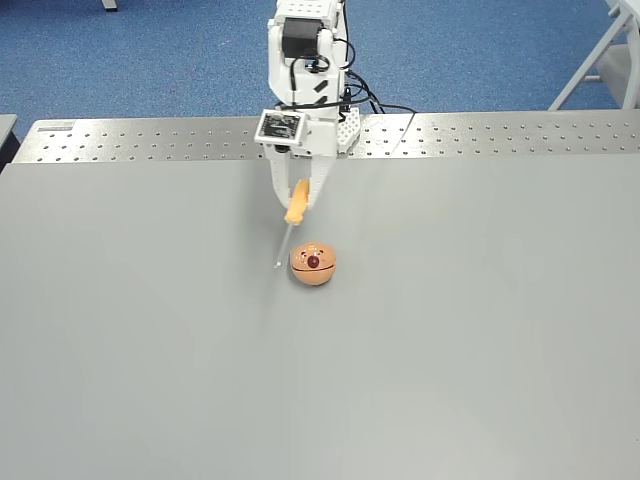} &
        \includegraphics[width=.32\linewidth]{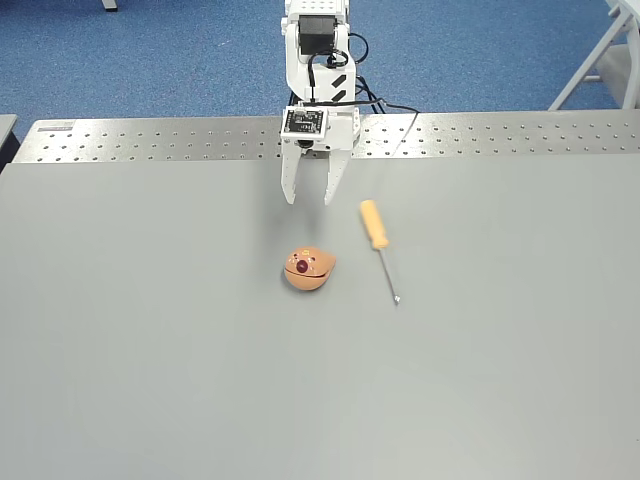} \\
        \multicolumn{3}{c}{Piper / Water bottle} \\
        \includegraphics[width=.32\linewidth]{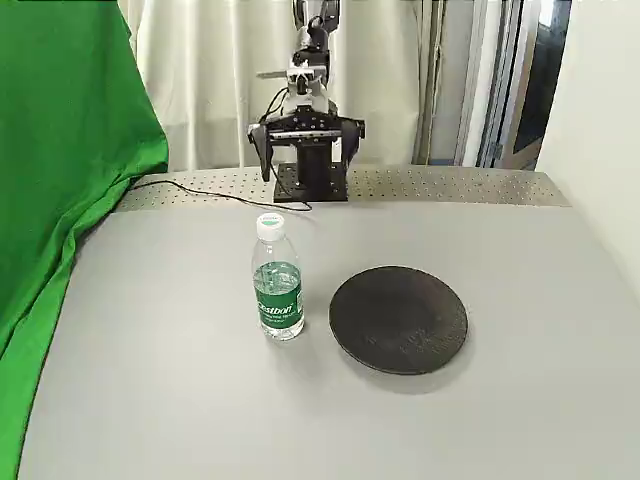} &
        \includegraphics[width=.32\linewidth]{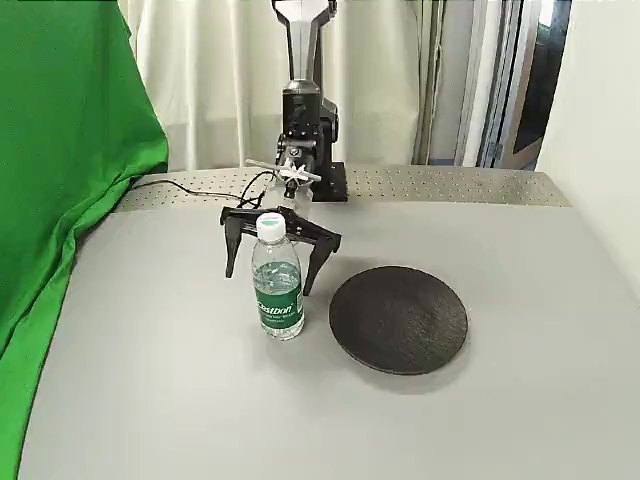} &
        \includegraphics[width=.32\linewidth]{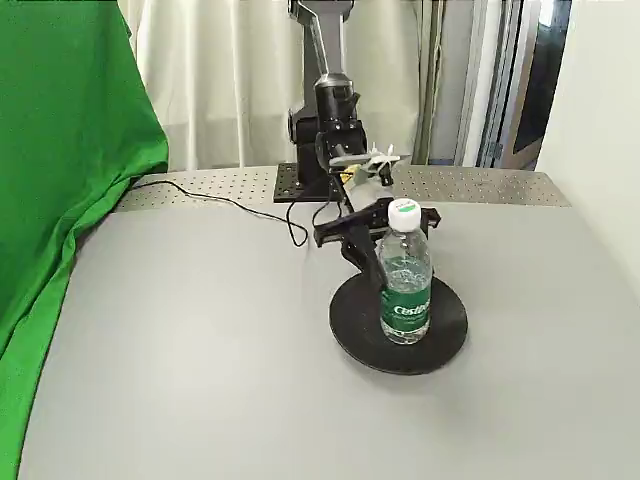} \\
        \multicolumn{3}{c}{Piper / Pour beer} \\
        \includegraphics[width=.32\linewidth]{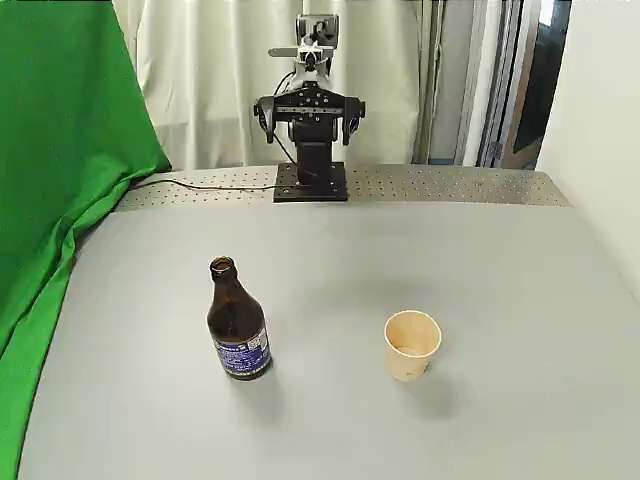} &
        \includegraphics[width=.32\linewidth]{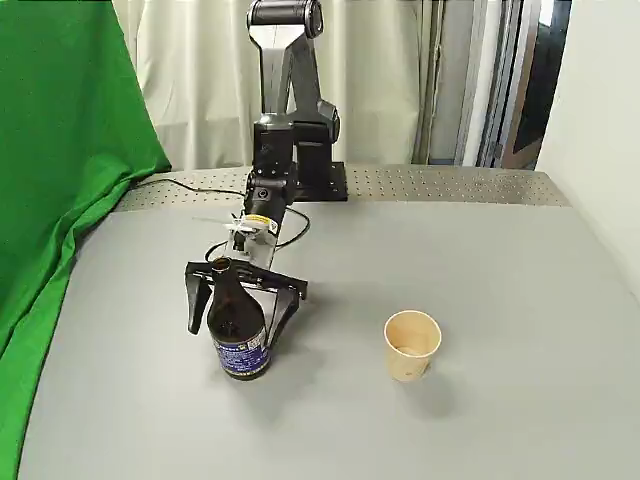} &
        \includegraphics[width=.32\linewidth]{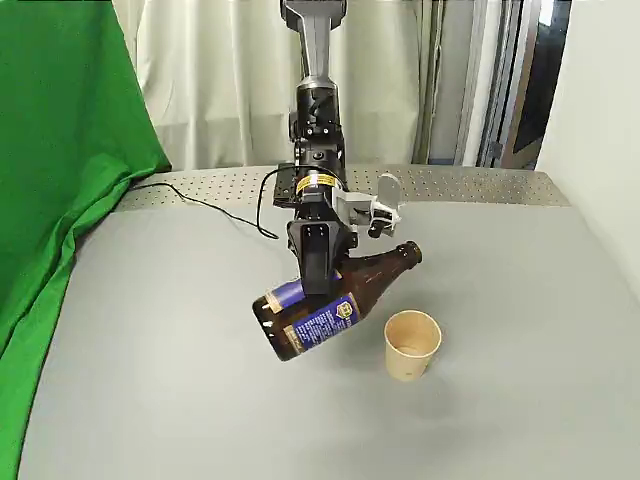} \\
        \multicolumn{3}{c}{Piper / Wooden fish} \\
        \includegraphics[width=.32\linewidth]{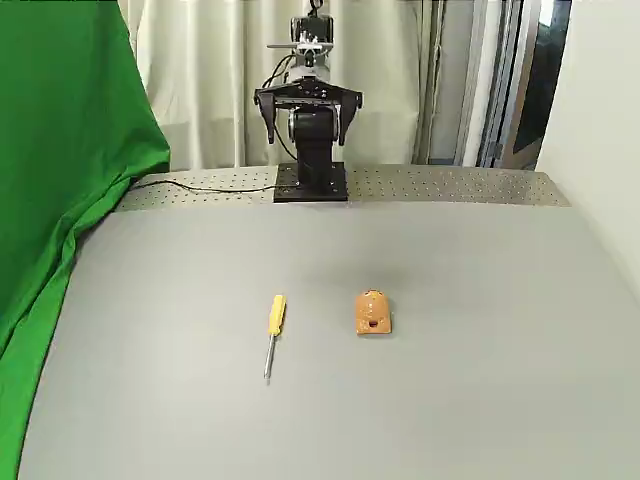} &
        \includegraphics[width=.32\linewidth]{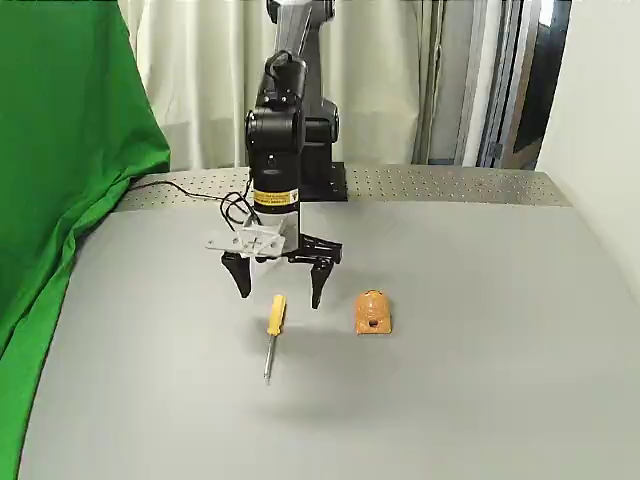} &
        \includegraphics[width=.32\linewidth]{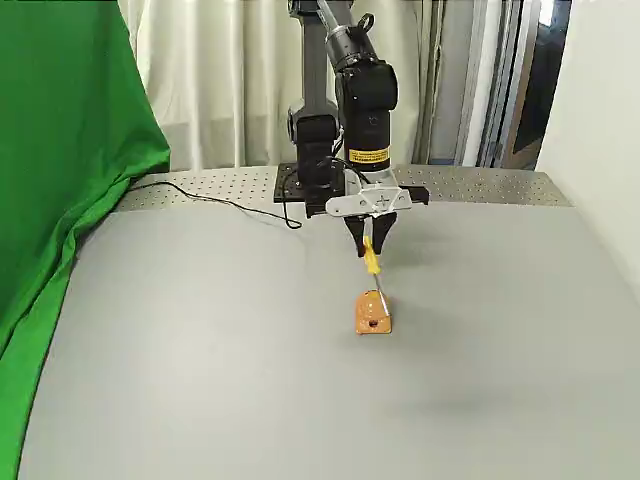} \\
    \end{tabular}
    \caption{Demonstration sequences for the three real robot tasks on SO-101 and single-arm Piper. Each row shows front camera observations at 15\%, 50\%, and 85\% of one demonstration, ordered from left to right. These examples illustrate the recorded trajectories and do not indicate evaluation success.}
    \label{fig:real-robot-examples}
\end{figure}

\end{document}